\documentclass[review]{elsarticle}
\usepackage[numbers]{natbib}
\usepackage{float}
\usepackage{ulem}
\usepackage{graphicx}
\usepackage{url}

\usepackage{array, caption, threeparttable}
\usepackage{algpseudocode}  
\usepackage{amsmath}  
\usepackage{algorithm}

\usepackage{algorithmicx}
\usepackage{arydshln}

\usepackage{subcaption}

\usepackage{xcolor}
\usepackage{stfloats}
\usepackage{tcolorbox}
\usepackage{amsmath}
\usepackage{amsfonts}
\usepackage{multirow}
\usepackage{booktabs}
\usepackage{listings}
\usepackage{pifont}
\usepackage{makecell}
\usepackage{nicematrix}
\usepackage{bm}
\usepackage{threeparttable}

\usepackage{pgfplots}
\usepackage{hyperref}
\hypersetup{hypertex=true,
colorlinks=true,
linkcolor=blue,
anchorcolor=blue,
citecolor=blue}

\definecolor{verylightgray}{rgb}{.97,.97,.97}

\newcommand{\yg}[1]{\textcolor{black}{#1}}
\newcommand{\gcy}[1]{\textcolor{black}{#1}}
\newcommand{\tool}{PSO-KDVA}

\algnewcommand\algorithmicinitialize{\textbf{Initialize}}
\algnewcommand\Initialize{\item[\algorithmicinitialize]}
\algnewcommand\algorithmicFor{\textbf{For}}
\algnewcommand\for{\item[\algorithmicFor]}

\journal{}

\begin{document}

\captionsetup[table]{
  labelfont=bf,
  labelsep=newline,
  singlelinecheck=false,
}

\begin{frontmatter}
	
    \title{Resource-Efficient Automatic Software Vulnerability Assessment via Knowledge Distillation and Particle Swarm Optimization}

    \author[NTU]{Chaoyang Gao}
    \ead{gcyol@outlook.com}
       
    \author[NTU,NJU]{Xiang Chen\corref{mycorrespondingauthor}}
    \ead{xchencs@ntu.edu.cn}
     
    \cortext[mycorrespondingauthor]{Corresponding author}

    \author[NTU]{Jiyu Wang}
    \ead{jyu.wang@outlook.com}

    \author[NTU]{Jibin Wang}
    \ead{W20040830@outlook.com}

    \author[NUAA,SMU]{Guang Yang}
    \ead{novelyg@outlook.com}

    \address[NTU]{School of Artificial Intelligence and Computer Science, Nantong University, Nantong, China}
    \address[NJU]{State Key Lab. for Novel Software Technology, Nanjing University, Nanjing, China}
    \address[NUAA]{College of Computer Science and Technology, Nanjing University of Aeronautics and Astronautics, Nanjing, China}
    \address[SMU]{School of Computing and Information Systems, Singapore Management University, Singapore}

\begin{abstract}
The increasing complexity of software systems has led to a surge in cybersecurity vulnerabilities, necessitating efficient and scalable solutions for vulnerability assessment. However, the deployment of large pre-trained models in real-world scenarios is hindered by their substantial computational and storage demands. To address this challenge, we propose a novel resource-efficient framework that integrates knowledge distillation and particle swarm optimization to enable automated vulnerability assessment.  
Our framework employs a two-stage approach: First, particle swarm optimization is utilized to optimize the architecture of a compact student model, balancing computational efficiency and model capacity. Second, knowledge distillation is applied to transfer critical vulnerability assessment knowledge from a large teacher model to the optimized student model. This process significantly reduces the model size while maintaining high performance.  
Experimental results on an enhanced MegaVul dataset, comprising 12,071 CVSS (Common Vulnerability Scoring System) v3 annotated vulnerabilities, demonstrate the effectiveness of our approach. Our approach achieves a 99.4\% reduction in model size while retaining 89.3\% of the original model's accuracy. Furthermore, it outperforms state-of-the-art baselines by 1.7\% in accuracy with 60\% fewer parameters. The framework also reduces training time by 72.1\% and architecture search time by 34.88\% compared to traditional genetic algorithms.

This study presents a novel framework for software vulnerability assessment in Cybersecurity, referred to as {\tool}, which leverages Particle Swarm Optimization (PSO) and Knowledge Distillation (KD) to create a lightweight model. With the increasing adoption of large pre-trained models in software engineering, their high computational and storage demands present significant challenges for real-time deployment, especially in resource-constrained environments. The proposed {\tool} approach addresses these challenges by optimizing the architecture of the student model through the PSO algorithm and transferring knowledge from the teacher model to the student model via knowledge distillation. This process significantly reduces both model size and inference time. Experimental results show that {\tool} maintains 89.3\% of the original model’s performance while compressing its size to just 0.6\% of the original. Furthermore, {\tool} outperforms existing models, such as the BiLSTM-based approach, by achieving a 1.7\% improvement in performance. The proposed framework demonstrates the feasibility of constructing high-performance, lightweight models for software vulnerability assessment, offering an effective solution for deployment in resource-constrained environments.

\end{abstract}


\end{frontmatter}

\section{Introduction}
\label{sec:intro}

Software vulnerabilities refer to defects or oversights in software systems that adversaries can exploit, potentially resulting in unauthorized data access, operational interruptions, financial damage, loss of trust, and compliance issues. Such vulnerabilities often stem from programming mistakes, flawed architecture, or improper configurations, and their exploitation can lead to serious consequences for organizations and individuals alike. To mitigate these risks, Software Vulnerability Assessment (SVA)~\cite{shah2015overview,le2022survey} is conducted to systematically detect, analyze, and rank potential security flaws within software applications. The main objective of SVA is to improve software security by proactively identifying and addressing threats, thus lowering the chances of successful attacks, meeting industry regulations, and fostering confidence among users and stakeholders. 

Current software vulnerability assessment tasks primarily rely on pre-trained large language models of code~\cite{le2019automated}. These models, trained on massive code datasets, excel in downstream software engineering tasks, such as code generation~\cite{yang2023exploitgen,zhou2023codebertscore} and vulnerability detection~\cite{hanif2022vulberta,wu2021peculiar}. However, these large models come with an enormous number of parameters; for example, models like CodeBERT have 125 million parameters and exceed 450 MB in size~\cite{feng2020codebert}, which brings notable challenges. Firstly, excessive parameters result in slow inference speed. In practical applications, inference with large models often requires a significant amount of time, especially on standard consumer-grade devices, where response delays exceed one second, far from meeting the demands of real-time, high-speed tasks~\cite{aye2020sequence}. Secondly, the model's large size complicates deployment. Storage requirements for large pre-trained code models frequently exceed the capacities of standard development environments or low-end devices. This is especially true in environments with limited storage, such as integrated development environments (IDEs) or embedded systems, where deploying these models is challenging~\cite{svyatkovskiy2021fast}. Additionally, large models typically demand greater memory and computational resources, restricting their application on resource-constrained devices.

More critically, the intensive computation requirements of these large models result in substantial energy consumption and carbon emissions, negatively impacting environmental sustainability~\cite{wei2023towards,shi2024efficient}. This trend is especially concerning as the use of large language models in software engineering tasks (such as program repair~\cite{jin2023inferfix,peng2024domain} and test generation~\cite{schafer2023empirical,yuan2023no}) continues to grow, heightening environmental concerns. For instance, each model inference involves a large number of floating-point operations (FLOPs)~\cite{clark2020electra}, which consumes significant energy resources. When these models are run frequently across multiple devices, cumulative carbon emissions further increase. Consequently, optimizing and compressing these models to reduce energy consumption and carbon footprint, while preserving performance as much as possible, has become a key challenge in current research and application of large code models.

In recent years, many studies~\cite{choudhary2020comprehensive,cheng2018model,deng2020model} have begun exploring model compression techniques to reduce model size, enhance inference speed, and decrease computational resource consumption. The most common model compression methods include model pruning, model quantization, and knowledge distillation. Model pruning~\cite{liu2018rethinking,lin2020dynamic,jiang2022model}, reduces the number of parameters by setting certain parameters to zero or removing parts of the network structure, such as layers or attention heads. However, pruning techniques have limited effectiveness in compressing large pre-trained code models. Even with the removal of network layers, a large embedding table often remains, as in CodeBERT, where it accounts for around 150 MB~\cite{shi2023compressing}. This makes it difficult to compress the model to meet stringent storage constraints. Model quantization~\cite{polino2018model,fan2020training} converts model parameters from 32-bit floating-point numbers to lower-bit fixed-point numbers, such as 8-bit or even lower, thereby reducing storage requirements and memory usage. While quantization can decrease model size to some extent, its impact on inference speed is limited, particularly when running on CPUs~\cite{ganesh2021compressing}, where quantized models still require significant resources.

Knowledge distillation~\cite{park2019relational,gou2021knowledge} involves training a smaller student model to mimic the behavior of a large teacher model, achieving knowledge transfer and model compression. The teacher model’s outputs serve as ``soft labels" for the student model, allowing the retention of essential information from the original model in a smaller structure. This approach has notable advantages, as it can maintain performance close to the original model while achieving better inference efficiency when running on standard devices~\cite{zhao2022decoupled}.

Although knowledge distillation effectively compresses model size and maintains high performance, it still faces several key challenges. The first is the capacity gap. Large code models, such as CodeBERT, possess extensive parameters and complex network structures, enabling them to represent rich knowledge, which is often referred to as ``high model capacity". In contrast, small student models have shallower network structures and lower dimensions, resulting in limited capacity and difficulty in effectively capturing the extensive knowledge in the large model~\cite{gou2021knowledge}. This capacity gap can easily lead to performance loss during the distillation process.

The complexity of selecting the student model architecture is another significant challenge in knowledge distillation. Choosing a suitable student model architecture is crucial for effective distillation, as it requires retaining as much knowledge as possible within limited storage space. However, selecting the student model architecture is essentially a large combinatorial optimization problem~\cite{shi2024greening}, involving choices around network structures (e.g., CNN, LSTM) and various hyperparameters (e.g., number of layers and attention heads). This vast search space makes it computationally infeasible to train and validate all possible student model architectures individually. Therefore, practical implementations require efficient and straightforward evaluation metrics to filter and identify potentially optimal model architectures.

To address the capacity gap and architectural complexity challenges in knowledge distillation, we propose a particle swarm optimization-based student model architecture search approach, {\tool}. Similar to prior research~\cite{gao2021residual,shi2023compressing,shi2024greening}, {\tool} uses Giga floating-point operations (GFLOPs) as an evaluation metric for computational cost. The difference between the original model size and the compressed student model size is used as a metric for assessing the capacity gap. By combining these two metrics, we construct a fitness function that guides the particle swarm optimization algorithm to converge quickly in the large architectural space to find an optimal student model architecture that balances performance and efficiency. After obtaining the optimal architecture for the student model, we apply knowledge distillation techniques to guide the student model in learning the teacher model’s knowledge. Specifically, we input unlabeled data into the teacher model for training and obtain prediction results for each data. These unlabeled data, along with the teacher model’s predictions, are then used to train the student model. 

We evaluate the effectiveness of our {\tool} approach using a dataset that contains 12,071 vulnerability entries. This dataset is based on the megavul~\cite{ni2024megavul} dataset, which we have expanded and modified. Following prior research, we use CodeBERT as the large code model for software vulnerability assessment and apply knowledge distillation for model compression. Specifically, our proposed {\tool} approach compresses the model size to 0.6\% of the original size while maintaining 89.3\% of the original model's performance. Furthermore, we compare the {\tool} approach with the state-of-the-art $\text{BiLSTM}_{soft}$~\cite{tang2019distilling} approach. The results indicate that the compressed model size of our approach is only 40\% of that of the $\text{BiLSTM}_{soft}$ approach, while the performance of the $\text{BiLSTM}_{soft}$ approach is only 98.3\% of that of the {\tool} approach. Additionally, to illustrate the execution efficiency of the particle swarm optimization (PSO)~\cite{jain2018review,shami2022particle,prajapati2018particle} algorithm, we compare it with the genetic algorithm (GA)~\cite{lambora2019genetic,mirjalili2019genetic}. The result shows that the PSO algorithm reduces the time cost of the model architecture search process by 34.88\%. Finally, we analyze the time cost of the {\tool} approach, revealing that the training time required by our approach is only 27.9\% of that needed for the pre-trained code model.

\textbf{The novelty and contributions} of our study can be summarized as follows:

\begin{itemize}

    \item \textbf{Perspective.} We are the first to compress large code models used for software vulnerability assessment, employing knowledge distillation to reduce model size while preserving performance as much as possible.

    \item \textbf{Approach.} We introduce the {\tool} approach, which leverages the particle swarm optimization algorithm to efficiently identify optimal model architectures within a large parameter space that meets specific constraints.

    \item \textbf{Practical Evaluation.} We conduct a comprehensive evaluation of {\tool}, demonstrating that it significantly reduces model size and training time while maintaining high performance.

\end{itemize}

\textbf{Open Science.} 
To facilitate replication and further research, we share our dataset and source code on GitHub (\url{https://github.com/judeomg/PSO-KDVA}).

\textbf{Paper Organization.} 
Section~\ref{sec:Background} introduces the research background and the research challenges.
Section~\ref{sec:Framework} presents the implementation details of our proposed approach {\tool}.
Section~\ref{Experimental Setup} describes our experimental setup. 
Section~\ref{sec:Results} shows our experimental results and main findings. 
Section~\ref{sec:Discussion} discusses the impact of compressed model sizes, and analyzes potential threats to our study. 
Section~\ref{sec:Related Work} summarizes related work. 
Section~\ref{sec:Conclusion} summarizes our study and discusses potential future directions.

\section{Background}
\label{sec:Background}

In this section, we first introduce the background of software vulnerability assessment. Then we introduce the model compression and knowledge distillation. Finally, we analyze the primary challenges faced in our research.

\begin{figure*}[!t]
	\centering
	\includegraphics[width=0.95\textwidth]{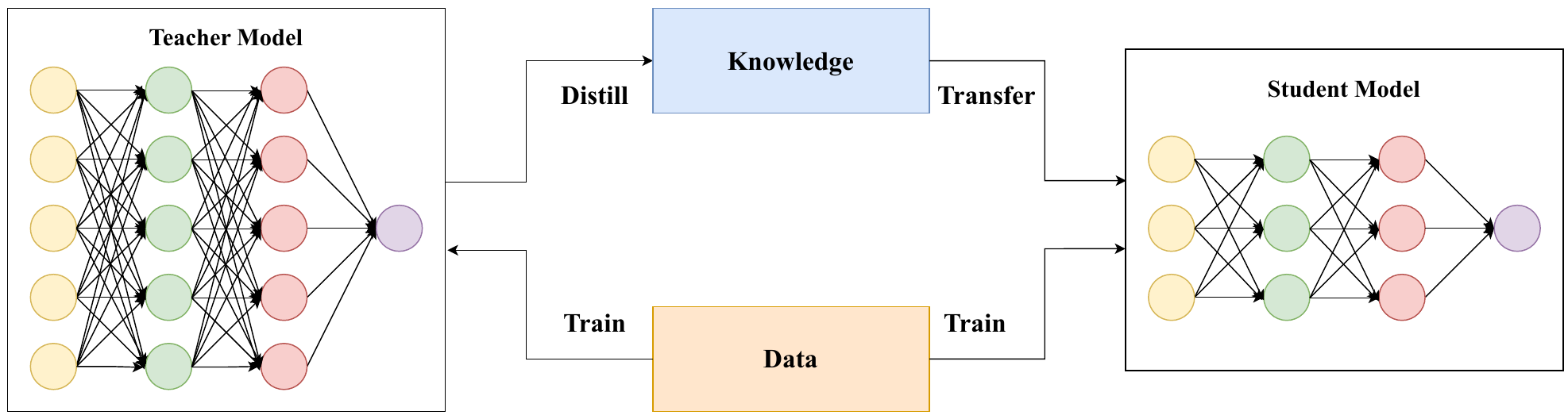}
	\caption{Knowledge distillation framework from teacher model to student model}
	\label{fig:knowledge_distillation}
\end{figure*}

\subsection{Software Vulnerability Assessment}

The core objective of software vulnerability assessment is to help developers and security experts identify and understand potential risks associated with vulnerabilities more quickly, allowing them to prioritize the most threatening issues under limited resources~\cite{le2022survey,du2019leopard}. With the rapid growth of cybersecurity threats, the number of reported vulnerabilities rises each year, making it essential to effectively evaluate the severity and impact of these vulnerabilities to optimize management and establish repair priorities~\cite{humayun2022security}.

\yg{\textbf{Formal Problem Definition.} Let $\mathcal{V} = \{v_1, v_2, \ldots, v_n\}$ be a set of software vulnerabilities, where each vulnerability $v_i$ is characterized by its code context $\mathbf{c}_i$. 
The vulnerability assessment task aims to learn a mapping function $f: \mathcal{C} \rightarrow \mathcal{S}$, where $\mathcal{C}$ represents the space of code contexts and $\mathcal{S}$ denotes the severity space. 
For automated vulnerability assessment, this can be formulated as:}

\yg{
\begin{equation}
\hat{s}_i = f(\mathbf{c}_i; \boldsymbol{\theta})
\end{equation}
}
\yg{where $\hat{s}_i$ is the predicted severity for vulnerability $v_i$, $\mathbf{c}_i$ is the encoded representation of the code context, and $\boldsymbol{\theta}$ represents the model parameters.}

To achieve standardized vulnerability assessment, the Common Vulnerability Scoring System (CVSS)~\cite{cvss} has become a widely used tool in the industry. CVSS provides quantifiable metrics for various characteristics of a vulnerability, including impact and exploitability, among others. These scores not only assist developers in understanding the potential security impact on systems but also guide them in allocating resources wisely to prioritize the remediation of vulnerabilities with the most significant impact.

Traditionally, CVSS scoring relies on manual evaluation by experts, but this approach faces significant delays~\cite{holm2015expert,beck2016using}. The manual scoring process can take days or even months to be released after detection and remediation, leading to a lag in response. To address this issue, automated vulnerability assessment methods have been proposed and increasingly adopted, aiming to resolve the delays caused by manual evaluation. These automated methods leverage machine learning and deep learning models to analyze and predict CVSS scores directly from code or commit records~\cite{liu2019vulnerability}. Existing automated methods often use pre-trained large code models to predict vulnerability severity~\cite{le2022survey,babalau2021severity,ganesh2021predicting,purba2023software}, which reduces the manual effort but increases computational costs. Therefore, we compress large code models using knowledge distillation, significantly reducing model size while maximizing performance. Our approach contributes to improved resource efficiency and environmental sustainability in automated vulnerability assessment.

\subsection{Model Compression and Knowledge Distillation}

Pre-trained large code models, such as CodeBERT, excel in tasks like code generation~\cite{khanfir2022codebert,zhou2023codebertscore}, clone detection~\cite{sonnekalb2022generalizability,khajezade2024investigating}, and vulnerability detection~\cite{nguyen2022regvd,zhou2024large}. 
However, their massive parameter sizes result in slow inference speeds, high memory demands, and substantial resource consumption. 
These large models not only restrict deployment on consumer-grade devices but also incur high computational and energy costs. 
Therefore, researchers are exploring various model compression techniques, such as model pruning~\cite{liu2018rethinking,lin2020dynamic,jiang2022model}, model quantization~\cite{polino2018model,fan2020training}, and knowledge distillation~\cite{park2019relational,gou2021knowledge} to reduce model size and improve inference efficiency.

\yg{\textbf{Model Compression Objective.} Given a teacher model $f_T$ with parameters $\boldsymbol{\theta}_T$ and $|\boldsymbol{\theta}_T|$ parameters, the goal of model compression is to obtain a student model $f_S$ with parameters $\boldsymbol{\theta}_S$ such that $|\boldsymbol{\theta}_S| \ll |\boldsymbol{\theta}_T|$ while maintaining comparable performance. The compression ratio can be defined as:}

\yg{
\begin{equation}
\rho = \frac{|\boldsymbol{\theta}_S|}{|\boldsymbol{\theta}_T|}
\end{equation}
}
\yg{Notice $\rho \ll 1$ indicates a significant compression.}

\textbf{Knowledge Distillation.} Knowledge distillation, an effective model compression technique, transfers knowledge from a large model (teacher model) to a smaller model (student model), significantly reducing model size while maintaining optimal performance. Figure~\ref{fig:knowledge_distillation} illustrates the framework of knowledge distillation. 
In the distillation process, the teacher model’s output serves as ``soft labels" for the student model, guiding it to learn the teacher model's behavior. This approach is well-suited for running on standard devices, effectively reducing inference time and resource consumption. Knowledge distillation has been applied to various tasks, such as image classification~\cite{mukherjee2019cogni,peng2019few}, recommendation systems~\cite{pan2019novel,chen2018adversarial}, and speech recognition~\cite{bai2019learn,ng2018teacher}. Knowledge distillation can be classified into task-specific distillation and task-agnostic distillation~\cite{gou2021knowledge}. Task-agnostic distillation~\cite{wang2020minilm,liang2023homodistil} is not limited to specific tasks but instead transfers broad knowledge from a general teacher model to the student model. While this approach enhances model versatility, it demands substantial computational resources and may underperform on certain tasks since it lacks task-specific optimization. In contrast, task-specific distillation~\cite{tang2019distilling,li2020knowledge} involves training or fine-tuning the teacher model on specific task data to generate more accurate predictions or feature representations, which are then transferred to the student model. This approach is less resource-intensive, and the student model typically achieves strong performance, as the knowledge it learns is tailored to the specific task. Therefore, our approach adopts task-specific distillation.

\yg{Formally, the fundamental principle of knowledge distillation lies in the soft target distribution provided by the teacher model. Given an input $\mathbf{x}$, the teacher model generates a probability distribution over classes:}

\yg{
\begin{equation}
p_i^T = \frac{\exp(z_i^T/\tau)}{\sum_j \exp(z_j^T/\tau)}
\end{equation}
}
\yg{where $z_i^T$ is the $i$-th logit output by the teacher model, and $\tau$ is the temperature parameter that controls the softness of the probability distribution. Similarly, the student model generates a probability distribution over classes:}

\yg{
\begin{equation}
p_i^S = \frac{\exp(z_i^S/\tau)}{\sum_j \exp(z_j^S/\tau)}
\end{equation}
}

\yg{The knowledge distillation loss combines the traditional cross-entropy loss with the distillation loss:}

\yg{
\begin{equation}
\mathcal{L}_{\text{KD}} = \alpha \mathcal{L}_{\text{CE}}(y, p^S) + (1-\alpha) \tau^2 \mathcal{L}_{\text{KL}}(p^T, p^S)
\end{equation}
}
\yg{where $\alpha$ denotes a weighting factor, $\mathcal{L}_{\text{CE}}$ is the cross-entropy loss with ground truth labels $y$, and $\mathcal{L}_{\text{KL}}$ is the Kullback-Leibler divergence loss:}

\yg{
\begin{equation}
\mathcal{L}_{\text{KL}}(p^T, p^S) = \sum_i p_i^T \log \frac{p_i^T}{p_i^S}
\end{equation}
}	

\yg{
\textbf{Theoretical Foundation of Knowledge Distillation.} The effectiveness of knowledge distillation can be understood from the following two theoretical perspectives:}

\yg{
\textbf{(1) Information Transfer.} The soft targets from the teacher model contain richer information than hard labels. The relative probabilities between non-target classes provide valuable information to help the student model generalize better.}

\yg{
\textbf{(2) Regularization Effect.} The distillation loss can act as a regularizer that prevents the student model from overfitting to the training data. This can be formalized as:}

\yg{
\begin{equation}
\mathcal{R}(f_S) = \mathbb{E}_{\mathbf{x} \sim \mathcal{D}} [D_{\text{KL}}(f_T(\mathbf{x}) \| f_S(\mathbf{x}))]
\end{equation}
}
\yg{where $\mathcal{R}(f_S)$ represents the regularization term that encourages the student to mimic the teacher's behavior.}

\subsection{Research Challenges}

However, applying knowledge distillation techniques to software vulnerability assessment presents several key challenges. First, due to the broad parameter space of models, selecting an appropriate student model architecture is complex, and finding a suitable architecture within an acceptable timeframe is crucial. Additionally, designing an effective optimization objective is essential so that the distilled student model not only reduces in size but also maintains high performance. Finally, analyzing the impact of model compression on performance is critical for evaluating the effectiveness of this approach.

\section{Approach}
\label{sec:Framework}

\begin{figure*}
	\centering
	\includegraphics[width=0.95\textwidth]{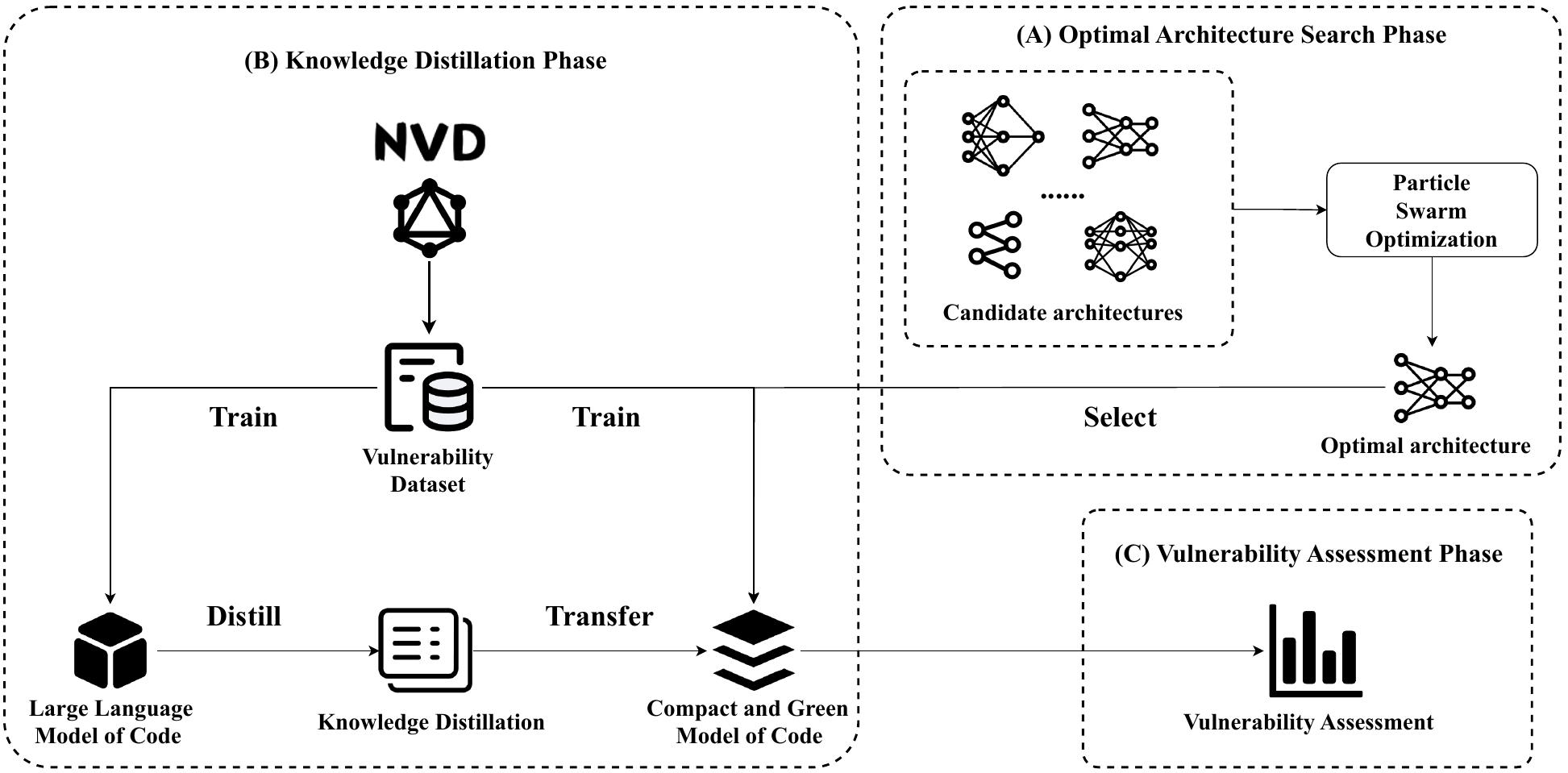}
	\caption{Framework of our proposed approach {\tool}}
	\label{fig:methodology}
\end{figure*}

Figure~\ref{fig:methodology} shows the framework of our proposed approach {\tool}. It is divided into three main phases. Specifically, in the \textbf{optimal architecture search phase}, the entire parameter configuration space is first determined based on the compressed model size. All possible combinations in this parameter space form the candidate architecture set for the student model. Then, the particle swarm optimization algorithm is applied to find an optimal student model architecture that meets the constraint conditions. In the \textbf{knowledge distillation phase}, a large code model is trained on a collected vulnerability dataset to acquire knowledge about software vulnerability assessment. After training, knowledge distillation is performed on the teacher model, transferring its knowledge to the optimal student model architecture to create a compact and efficient student model. Finally, in the \textbf{vulnerability assessment phase}, the distilled student model is tested on the vulnerability assessment task to evaluate its capability in predicting vulnerability severity on a real dataset. In the remainder of this section, we present the implementation details of our {\tool} approach.

\subsection{Optimal Architecture Search Phase}
\label{sec3.1}

Knowledge distillation is a model compression technique that aims to transfer the knowledge of a teacher model (a large model) to a student model (a smaller model), thereby significantly reducing the model's size while retaining the performance of the larger model. The key to this approach lies in identifying the optimal architecture for the student model to ensure it can effectively learn the teacher model's knowledge and achieve good performance on specific tasks~\cite{cho2019efficacy}. Balancing model size reduction with maintaining performance is essentially a combinatorial optimization problem. Searching for the most suitable student model architecture in the parameter space is inherently a search problem: adjusting given objective functions and constraints to find the optimal solution within an acceptable time frame~\cite{shi2024greening}. To address this, we must overcome three key challenges.

First, selecting an appropriate architecture for the student model requires searching through numerous configuration combinations to find the one that optimizes the smaller model’s structure while maintaining performance. This process is computationally intensive and time-consuming. To address this issue, we propose the {\tool} approach, which employs the particle swarm optimization algorithm to search the configuration space of the model. PSO is a global optimization algorithm based on swarm intelligence, which simulates the collective behavior of birds to locate the optimal solution~\cite{kennedy1995particle}. In the PSO algorithm, each candidate solution is treated as a ``particle" that ``flies" through the configuration space, gradually adjusting its position to minimize the objective function value. A particle's position is updated based on its own historical best position (individual best) and the swarm’s overall best position (global best)~\cite{jain2018review,shami2022particle}. Through this iterative process, the particle swarm converges towards the optimal solution, allowing the identification of the optimal student model configuration within a limited time.

Second, an effective metric is required to comprehensively assess both the size and performance of the student model after knowledge distillation. Typically, each candidate model would need to undergo training and evaluation, which is resource-intensive and often unfeasible. Therefore, a simple and efficient predictive metric is needed to guide the model architecture search. Based on previous research~\cite{shi2024greening,gao2021residual,wu2021low}, we use Giga floating-point operations (GFLOPs) as a measure of computational cost.

Finally, because the teacher model usually possesses greater complexity and learning capability, the student model, constrained by its network structure, may struggle to fully absorb and replicate the knowledge of the larger model. This difference in model capacity may lead to suboptimal performance in the student model. To address this, we draw from similar research and employ soft labels as inputs for the student model~\cite{zhou2021rethinking,aguilar2020knowledge,zi2021revisiting}, using knowledge distillation to guide the student model’s learning of the teacher model's knowledge. This soft label approach helps the student model understand the reasoning and prediction process of the teacher model, thereby enhancing its performance.

\subsubsection{Hyperparameter Configuration Space}
\label{Configuration Space}

\newcolumntype{L}[1]{>{\raggedright\arraybackslash}p{#1}}
\newcolumntype{C}[1]{>{\centering\arraybackslash}p{#1}}

\newcolumntype{R}[1]{>{\raggedleft\arraybackslash}p{#1}}
\begin{table*}[htb]

\centering
\caption{The hyperparameters included in the configuration space and their search ranges, as well as the values of each hyperparameter in the teacher model. The parameters in bold are those that can affect model size.}
\resizebox{12cm}{!}
{
\begin{tabular}{{L{5cm}L{4cm}L{6cm}}}

\toprule
\textbf{Hyperparameter Name} & \textbf{Pre-trained Models} & \textbf{Search Space} \\

\midrule
\emph{tokenizer} & ``Byte-Pair Encoding" & [``Byte-Pair Encoding"] \\

\hline
\textbf{\emph{vocab\_size}} & 50265 & range (1000, 50265), interval = 1000 \\

\hline
\textbf{\emph{num\_hidden\_layers}} & 12 & range (1, 12), interval = 1 \\

\hline
\textbf{\emph{hidden\_size}} & 768 & range (16, 768), interval = 16 \\

\hline
\emph{hidden\_act} & ``GELU" & [``GELU"] \\

\hline
\emph{hidden\_dropout\_prob} & 0.1 & [0.1] \\

\hline
\textbf{\emph{intermediate\_size}} & 3072 & range (16, 3072), interval = 32 \\

\hline
\textbf{\emph{num\_attention\_heads}} & 12 & range (1, 12), interval = 1 \\

\hline
\emph{attention\_probs\_dropout\_prob} & 0.1 & [0.1] \\

\hline
\emph{max\_sequence\_length} & 512 & [512] \\

\hline
\emph{position\_embedding\_type} & ``absolute" & [``absolute"] \\

\hline
\emph{learning\_rate} & 5e-5 & [1e-5, 5e-5, 1e-4, 2e-4, 5e-4, 1e-3, 2e-3] \\

\hline
\emph{batch\_size} & 32 & [32] \\

\bottomrule
\end{tabular}
}
\label{tab1}
\end{table*}

In this section, we define the hyperparameter search space for training the student model. This search space includes multiple hyperparameters related to model architecture and training configuration~\cite{shi2024greening,liu2020fastbert}. Each hyperparameter's range is carefully designed to ensure the effectiveness of the student model and to control the size of the search space. Table~\ref{tab1} lists these hyperparameters along with their respective ranges, as well as the values assigned to each hyperparameter in the teacher model. A detailed description follows.

\emph{Tokenizer}: Converts input text into tokens understandable by large code models, supporting four choices—``Byte-Pair Encoding"~\cite{vilar2021statistical}, ``WordPiece"~\cite{pan2021empirical}, ``Unigram"~\cite{kudo2018subword} and ``Word"~\cite{karampatsis2020big}. The pretrained model uses ``Byte-Pair Encoding". Since this parameter does not affect model size and only impacts model performance, we use the same settings as the fine-tuned teacher model in the student model architecture to achieve optimal performance. For parameters not shown in bold in the table, we maintain the same settings as the fine-tuned teacher model.

\emph{Vocab\_size}: Defines the size of the vocabulary, with a range of 1,000 to 50,265 and a step size of 1,000. A smaller vocabulary can effectively compress the model, but it must retain sufficient expressiveness.

\emph{Num\_hidden\_layers}: The number of hidden layers, controlling the model's depth, with a range of 1 to 12 and a step of 1. Reducing layers helps decrease model size but requires balancing performance and model capacity.

\emph{Hidden\_size}: The number of hidden units per layer, ranging from 16 to 768 with a step size of 16. More hidden units increase the model's representational power but also its size and computational cost.

\emph{Hidden\_act}: The type of activation function, with choices of ``GELU"~\cite{jiang2021treebert}, ``ReLU"~\cite{ishtiaq2021bert2code}, ``SiLU"~\cite{elfwing2018sigmoid} and ``GELU\_new"~\cite{shi2024greening}. 
The fine-tuned teacher model and the student model both set this parameter to ``GELU".

\emph{Hidden\_dropout\_prob}~\cite{mashhadi2021applying}: Dropout probability for hidden layers to prevent overfitting. The fine-tuned teacher model and the student model both set this parameter to 0.1.

\emph{Intermediate\_size}~\cite{mashhadi2021applying}: The size of the intermediate layer in the feedforward network, ranging from 16 to 3,072 with a step of 32. A larger size increases model capacity but also computational overhead.

\emph{Num\_attention\_heads}: The number of attention heads, representing how many features the model attends to in each layer. The search range includes 1, 2, 4, and 8, with the pretrained model using 12 heads.

\emph{Attention\_probs\_dropout\_prob}~\cite{zhou2021assessing}: Dropout probability for attention scores. The fine-tuned teacher model and the student model both set this parameter to 0.1.

\emph{Max\_sequence\_length}: The maximum input sequence length, allowing for handling of texts with different lengths. The fine-tuned teacher model and the student model both set this parameter to 512.

\emph{Position\_embedding\_type}~\cite{sharma2022exploratory}: The type of positional encoding scheme, including ``absolute", ``relative\_key" and ``relative\_key\_query". The fine-tuned teacher model and the student model both set this parameter to ``absolute".

\emph{Learning\_rate}: The learning rate adjusts to accommodate different training speeds. The fine-tuned teacher model sets this parameter to 5e-5, while the student model sets it to 5e-4.

\emph{Batch\_size}: Batch size per training iteration. The fine-tuned teacher model and the student model both set this parameter to 32.

The search ranges for these hyperparameters are designed not to exceed the search limits provided in Table~\ref{tab1}, ensuring that model simplification does not excessively impact its performance. Furthermore, by setting specific adjustment intervals for each hyperparameter, the configuration avoids the creation of potentially invalid model structures. Conducting the search within this configuration space allows for the model’s size to be compressed while maximizing its expressiveness and performance.

\subsubsection{PSO-based Search}

Based on the hyperparameter ranges analyzed in Section~\ref{Configuration Space} for the configuration space, there are approximately \(9.2 \times 10^{13}\) valid student model architectures. However, training and validating all possible architectures is clearly impractical. Therefore, we formulate the evaluation approach as an optimization problem, aiming to find the optimal solution for the objective function within given constraints. Our proposed {\tool} approach employs the PSO algorithm to search for the optimal model architecture that meets these constraints. Algorithm~\ref{algo_pso} illustrates the pseudocode for the PSO algorithm we use. Specifically, the process begins with randomly initializing the particles' position and velocity information, ensuring they are distributed within the hyperparameter range $c$ for model architecture (Line 1). We then initialize each particle's individual best solution (Line 2 to 4) and the global best solution (Line 5). The algorithm then enters an iterative loop where each particle's position and velocity are updated. The new velocity is calculated based on the particle’s current velocity, individual best position, and global best position. Using this updated velocity, each particle adjusts its position, moving to a new location (Lines 6 to 17), followed by calculating and updating the value of the fitness function (Lines 10 to 15). Finally, the model returns the solution with the highest fitness function value (Lines 18 to 19).

\begin{algorithm}[!t]
\caption{PSO Algorithm Pseudo-Code}
\label{algo_pso}
\begin{algorithmic}[1]
\Statex \textbf{Input:} $c$: value ranges of architecture-related hyperparameters, 
       $w$: inertia weight, 
       $c_1, c_2$: acceleration coefficients, 
       $maxIter$: maximum number of iterations, 
       $swarmSize$: number of particles in the swarm
\Statex \textbf{Output:} $c'$: optimized architecture-related hyperparameters

\State Initialize swarm with random positions and velocities within $c$
\For{each particle $i$ in swarm}
    \State $pBest[i] \gets particle[i]$  \Comment{individual best}
\EndFor
\State $gBest \gets \arg\max(pBest)$  \Comment{global best}

\For{$iter = 0$ to $maxIter$}
    \For{each particle $i$ in swarm}
        
        \State $velocity[i+1] \gets w \cdot velocity[i]$
        \Statex \hspace{3em} $+ c_1 \cdot \text{random()} \cdot (pBest[i] - position[i])$
        \Statex \hspace{3em} $+ c_2 \cdot \text{random()} \cdot (gBest - position[i])$

        \State $position[i+1] \gets position[i] + velocity[i+1]$

        \If{$fitness(position[i]) > fitness(pBest[i])$}
            \State $pBest[i] \gets position[i]$
        \EndIf

        \If{$fitness(pBest[i]) > fitness(gBest)$}
            \State $gBest \gets pBest[i]$
        \EndIf
    \EndFor
\EndFor
\State $c' \gets gBest$
\State \Return $c'$

\end{algorithmic}
\end{algorithm}

\textbf{Particle Representation}: In the PSO algorithm, a particle represents a feasible solution to the target problem~\cite{jain2018review,shami2022particle}, defined by a set of hyperparameters with specific values. In our study, each particle’s position represents a candidate configuration of hyperparameters, where each dimension of the position vector corresponds to a specific hyperparameter and its current value. The particle stores these hyperparameter values in a vector structure. The search space that particles move through consists of all hyperparameters, forming a 13-dimensional space. For example, a particle can be represented as [``Byte-Pair Encoding", 3000, 12, 96, ``GELU", 0.1, 3072, 12, 0.1, 512, ``absolute", 5e-5, 32]. The values in this vector are derived from the configuration space outlined in Section~\ref{Configuration Space}. Drawing from previous studies~\cite{prajapati2018particle,yang2021software}, we initialize a swarm of particles, with each hyperparameter's position randomly set within its predefined range. Each particle also has a specific velocity, which directs its movement within the search space. The randomly initialized particles collectively form the swarm, serving as the starting point for the iterative optimization process. 
\gcy{During the search process, the PSO algorithm updates the position of each particle iteratively. Each particle's movement is influenced by its own previous best position and the best position found by the entire swarm. The algorithm uses a set of parameters to control this movement. The inertia weight determines how much the particle's current velocity affects its next movement. Acceleration coefficients, along with random values, determine how strongly the particle is attracted to its personal best and the global best positions. In each iteration, all particles adjust their positions based on their individual best positions and the global best position identified by the swarm. Through this process, particles gradually converge toward the optimal solution.}

\textbf{Fitness Function}: In the PSO algorithm, the fitness function evaluates the quality of each particle's current position in the search space. A higher fitness value indicates that the particle's configuration is closer to meeting the model's design constraints. The objective of the {\tool} approach is to compress the model size while retaining as much of the model's actual performance as possible. Thus, the fitness function must account for both factors.

For evaluating model performance, we adopt Giga Floating-Point Operations (GFLOPs) as the metric, based on prior research~\cite{gao2021residual,shi2023compressing}. Since floating-point operations are more complex for computers to handle than integer operations, GFLOPs are commonly used to measure the complexity of computational tasks. Higher GFLOPs indicate a greater computational load, which generally suggests a more complex model with stronger computational capability and the ability to process more information and features, leading to better model performance.

\gcy{We choose GFLOPs over other common metrics like FLOPs or latency for the following reasons. First, the numerical range of GFLOPs (where GFLOPs is defined as FLOPs divided by $10^{9}$) in our experiments is well-suited for comparison with model size. We observe that GFLOPs values typically range from 1 to 3, while the model size difference is in the range of 0 to 1. This compatibility in numerical range makes it easier to combine these two metrics in the fitness function. Second, latency can vary greatly depending on the specific hardware platform and implementation environment. This variability makes it challenging to use latency as a consistent metric for optimizing models across different experimental setups. In contrast, GFLOPs provides a hardware-agnostic measure of computational complexity, directly reflects the number of operations, and is more relevant to resource constraints.}

For model size evaluation, we use the difference in capacity between the compressed student model and the original teacher model. Since the size of the teacher model remains constant, a larger difference implies a smaller, more compressed student model.

To achieve both objectives—compressing the model size while maximizing model performance—the PSO algorithm's fitness function is designed as follows:
\begin{equation}
    Fitness(i)=GFLOPs-|S-s_i|
\end{equation}
where $i$ represents the $i$-th particle, $GFLOPs$ indicates the computational capability of the student model, $S$ denotes the size of the teacher model, and $s_i$ represents the size of the student model under the current hyperparameter configuration.

\subsection{Knowledge Distillation Phase}

\begin{algorithm}[!t]
\caption{Knowledge Distillation Algorithm Pseudo-Code}
\label{algo_kd}
\begin{algorithmic}[1]
\Statex \textbf{Input:} $L$: teacher model, 
       $D$: training dataset, 
       $T$: temperature parameter
\Statex \textbf{Output:} $S$: student model

\For{$d$ in $D$}
    \State $p, q = L(d), S(d)$
    \State $loss = \text{softmax}(\frac{p}{T}) * \log(\text{softmax}(\frac{q}{T})) * T^2$
    \State $S.\text{update}(loss)$
\EndFor

\State \Return $S$

\end{algorithmic}
\end{algorithm}

Knowledge distillation is a widely used model compression technique that reduces the space of large code models by transferring knowledge from a pretrained teacher model to a smaller student model. This process allows the student model to closely approximate the teacher model's performance while reducing model size~\cite{park2019relational,peng2024domain}. In our study, we employ a task-specific distillation method based on the approach proposed by Hinton et al.~\cite{hinton2015distilling}, which has proven effective for large code models~\cite{zhao2022decoupled,shi2023compressing}.

As described in Algorithm~\ref{algo_kd}, the distillation process begins by inputting the same training data $D$ into both the teacher model $L$ and the student model $S$. For each input $d$ in $D$, the teacher and student models produce output probabilities $p$ and $q$, respectively (Line 2). The objective is to minimize the difference between these outputs, making the student model’s behavior similar to that of the teacher model. The core of this approach lies in the loss function (Line 3), where a softened version of the softmax output is applied using a temperature parameter $T$. By smoothing the probability distribution, this parameter facilitates the transfer of knowledge from the teacher model to the student model. The student model is trained to minimize this loss, iteratively adjusting its parameters (Line 4) to align its outputs with those of the fixed teacher model.

A notable advantage of our approach is its reliance on unlabeled data for training, addressing the challenge of obtaining labeled datasets. Unlabeled data can be easily collected from open-source repositories, and this approach aligns with previous findings~\cite{radosavovic2018data,jiao2019tinybert,tang2019distilling}, which suggests that label-free knowledge distillation can achieve competitive results.

\subsection{Vulnerability Assessment Phase}

In the vulnerability assessment phase, we use the distilled student model to evaluate the severity levels of software vulnerabilities. To predict the severity of vulnerable code, the model processes the input code and assesses the vulnerability based on the learned code representations. Given the vast number of software vulnerabilities, evaluating their severity requires significant computational resources. The lightweight nature of knowledge distillation makes it particularly well-suited for this task, allowing for efficient evaluation in real-time or on resource-constrained devices while maintaining assessment accuracy.

\section{Experimental Setup}
\label{Experimental Setup}

In this section, we first introduce our research questions and their design motivation. 
Then, we provide detailed information about the experimental subject, baseline, performance metrics, and running platform.

\subsection{Research Questions}

To show the competitiveness of {\tool} and the rationale of the component settings in {\tool}, we design the following four research questions (RQs) in our study.

\textbf{RQ1: How much can the {\tool} approach compress the model?} 

\textbf{Motivation}:
Current large code models have an enormous number of parameters, which not only increases memory requirements and inference latency but also limits their applicability on resource-constrained devices. To enhance model usability and reduce the environmental impact of resource consumption, we propose the {\tool} approach, which aims to efficiently compress the model using particle swarm optimization and knowledge distillation techniques. Therefore, it is necessary to evaluate the actual performance of the {\tool} approach in terms of model size compression. To further demonstrate the effectiveness of our proposed approach, we also compare it with the state-of-the-art knowledge distillation technique, $\text{BiLSTM}_{soft}$~\cite{tang2019distilling}.

\textbf{RQ2: Can the {\tool} approach minimize the loss in accuracy for vulnerability assessment while compressing the model size?} 

\textbf{Motivation}:
Although large models excel in tasks such as code generation and vulnerability detection, model compression may hinder the student model’s ability to effectively capture the teacher model’s knowledge, leading to reduced prediction accuracy. Thus, knowledge distillation faces a trade-off between performance loss and model size reduction. This research question aims to evaluate whether the {\tool} approach can retain high model performance while compressing model size.

\textbf{RQ3: To what extent does model size compression affect accuracy?} 

\textbf{Motivation}:
Reducing model capacity may impact performance, especially in complex multi-class vulnerability assessment tasks where performance is directly linked to accuracy. To better understand the relationship between model size and performance, it is necessary to explore how varying levels of compression impact model accuracy. This will help identify an optimal balance between compression effectiveness and performance, providing guidance for future model optimization and practical applications.

\textbf{RQ4: What is the execution efficiency of the PSO algorithm?} 

\textbf{Motivation}:
The {\tool} approach requires finding the optimal student model architecture before performing knowledge distillation. The computational efficiency of this process directly affects the overall time cost of model compression. Therefore, analyzing the execution efficiency of the PSO algorithm is particularly crucial. To this end, we consider comparing it with another commonly used and effective search algorithm, the Genetic Algorithm (GA) ~\cite{lambora2019genetic,mirjalili2019genetic}.

\subsection{Experimental Subject}

Our study utilizes an enhanced version of the MegaVul dataset initially constructed by Ni et al.~\cite{ni2024megavul}, which contains 17,975 C/C++ vulnerability records. This dataset provides rich vulnerability information, including vulnerable code, descriptions, and severity levels. However, Ni et al.'s dataset relies on the CVSS v2~\cite{le2019automated,le2021deepcva} vulnerability scoring standard. Building on their work, we retrieved CVSS v3 scores for all vulnerabilities. Compared to CVSS v2, the enhanced structure of CVSS v3 more accurately reflects vulnerability risks, aligning well with our research objectives. We removed vulnerabilities that did not have retrievable CVSS v3 scores, resulting in a final dataset of 12,071 vulnerabilities scored using the CVSS v3 standard.

We used stratified sampling to divide the dataset into a training set (80\%), validation set (10\%), and test set (10\%), ensuring a consistent distribution of severity levels across all subsets. Table~\ref{tab2} presents detailed information about these subsets, specifying the number of vulnerabilities at each severity level in each subset. For example, the training set contains 4,454 high-severity vulnerabilities, while the validation and test sets each contain 577 high-severity vulnerabilities.

\begin{table}[htb]

\centering
\caption{Statistical information of our experimental subject.}
\resizebox{8.6cm}{!}
{
\begin{tabular}{lccc}
\toprule
\textbf{Statistic} & \textbf{\makecell[c]{Train}} & \textbf{\makecell[c]{Validation}} & \textbf{\makecell[c]{Test}} \\
\midrule
Number & 9,656 & 1,207 & 1,208 \\
Number of Critical severity & 1,169 & 146 & 147 \\
Number of High severity & 4,454 & 577 & 577 \\
Number of Medium severity & 3,795 & 474 & 475 \\
Number of Low severity & 238 & 30 & 29 \\
\bottomrule
\end{tabular}
}
\label{tab2} 
\end{table}

\subsection{Baseline}

To validate the effectiveness of our proposed {\tool} approach, we selected the state-of-the-art knowledge distillation approach, $\text{BiLSTM}_{soft}$~\cite{tang2019distilling}, as a baseline for comparison. Given that we are the first to apply knowledge distillation to software vulnerability assessment, there is limited related research, so we focus primarily on the $\text{BiLSTM}_{soft}$ approach. The $\text{BiLSTM}_{soft}$ approach, proposed by Tang et al., transfers task-specific knowledge from a large pretrained language model, BERT, to a simpler and lighter BiLSTM-based model. This method employs knowledge distillation by fine-tuning the BERT model (teacher) on specific NLP tasks and using its output (logits) to train the smaller BiLSTM model (student) to mimic the teacher model’s behavior.

The approach was evaluated on multiple tasks within the GLUE benchmark, demonstrating competitive results with far fewer parameters and faster inference times than BERT, making it suitable for real-time or resource-constrained applications. Additionally, the $\text{BiLSTM}_{soft}$ approach also uses the soft labels from the teacher model’s outputs to train the student model~\cite{tang2019distilling}, consistent with our experimental setup. For implementing the $\text{BiLSTM}_{soft}$ approach, we utilized their open-source repository and maintained the same settings for model hyperparameters and other configurations.

\gcy{Notice the $\text{BiLSTM}_{soft}$ approach represents the most effective approach in the field of vulnerability assessment. In the future, we aim to explore the application of compression techniques from other domains to further optimize the performance and adaptability of the model.}

\subsection{Performance Metrics}
\label{sec4.4}

To evaluate the performance of our proposed {\tool} approach in the vulnerability assessment task, as well as the compression capability of the knowledge distillation approach, we use four metrics: Model Size in Memory, Time Cost, Accuracy, and MCC (Matthews Correlation Coefficient). Among these, Time Cost and MCC will be discussed further in section~\ref{sec:Discussion}. These metrics are detailed as follows:

\begin{itemize}
    \item \textbf{Model Size in Memory}. This metric indicates the memory footprint of the model, typically measured in MB or GB~\cite{rajbhandari2020zero}. It reflects the storage requirements of the model, which is particularly relevant for resource-constrained devices or applications where smaller models are generally easier to deploy.

    \item \textbf{Time Cost}.
    The time required for the model to perform a specific task is usually in seconds or milliseconds. This metric evaluates the inference efficiency of the model, with lower time costs indicating faster responses, which is crucial for real-time or low-latency applications. 

    \item \textbf{Accuracy}.
    The accuracy of the model in the vulnerability assessment task represents the proportion of correct predictions. This metric directly reflects the model’s overall performance, with higher accuracy indicating a more precise assessment of vulnerability severity levels.

    \item \textbf{MCC}.
    This metric measures the quality of the model’s predictions across different classes, particularly when there is a class imbalance problem in the dataset~\cite{gorodkin2004comparing}. MCC provides a comprehensive evaluation standard, ranging from -1 to 1, where higher values indicate stronger classification ability, and 0 suggests performance equivalent to random classification.
\end{itemize}

Using these metrics, we can comprehensively assess the compression efficiency and performance loss of our {\tool} approach.

\subsection{Running Platform}

All experiments are conducted on a server equipped with a GeForce RTX 4090 GPU with 24GB of graphic memory, running the Windows 10 operating system. The program is theoretically capable of running without a GPU and can be adapted to operate on other operating systems, such as Linux.

\section{Experimental Results}
\label{sec:Results}

\subsection{RQ1: How much can the {\tool} approach compress the model?}

\textbf{Approach:}
To investigate the compression capability of our proposed {\tool} approach on the large code model (CodeBERT), we compare it with a state-of-the-art knowledge distillation baseline and evaluate the results. For {\tool}, we employ the model architecture search strategy described in Section~\ref{sec3.1}, using the PSO algorithm to identify the optimal student model configuration space. Compression performance evaluation is based on model size and storage space, following the evaluation metrics introduced in Section~\ref{sec4.4}.

\textbf{Results:} 
The comparison of compression capabilities between {\tool} and the baseline model is shown in Table~\ref{tab:rq1}. Our proposed {\tool} approach achieves efficient model compression, significantly reducing model size. Specifically, {\tool} compresses the model to 0.63\% of its original size, requiring only 3 MB of storage space. In contrast, $\text{BiLSTM}_{soft}$ reduces the model size to 1.58\% of the original, with a storage requirement of 7.5 MB. Further analysis shows that the {\tool} approach reduces model storage by 4.5 MB compared to $\text{BiLSTM}_{soft}$, a reduction of 60\% relative to the $\text{BiLSTM}_{soft}$ approach. This indicates that {\tool} outperforms $\text{BiLSTM}_{soft}$ in terms of compression efficiency.

\begin{table*}[t!]
    \centering
    \caption{Comparison of compression effectiveness and performance between the teacher model (CodeBERT) and student models ($\text{BiLSTM}_{soft}$ and {\tool}). The values in the ``Drop" row for the student models indicate their improvement ratio relative to the teacher model. In the last row of the table, the ``Drop" value shows the improvement ratio of the {\tool} approach over the $\text{BiLSTM}_{soft}$ baseline.}
    \normalfont
\resizebox{12cm}{!}
{
    
    \begin{tabular}{{ccccccc}}
    \toprule
    \multirow{2}{*}{Approach} & 
    \multicolumn{2}{c}{Capacity} &
    \multicolumn{2}{c}{Vulnerability Assessment} \\ 
    \cmidrule(l){2-7} 
     & Model Size (MB) & Drop (\%) & Accuracy (\%) & Drop (\%) & \\ \midrule
    CodeBERT & 476 & - & 60.93 & - \\
    \hdashline     $\text{BiLSTM}_{soft}$ & 7.5 &  
    98.4 & 53.48 & 12.23 \\
    \textbf{{\tool}} & \textbf{3} & \textbf{99.4} & \bf{54.39} &  \bf{10.73}  \\

    \bottomrule
    
    Improvement of {\tool} over $\text{BiLSTM}_{soft}$ & 4.5 & 60 & 0.91 & 1.70 \\
    
    \bottomrule
    \end{tabular}
    \label{tab:rq1}
}
\end{table*}

\begin{tcolorbox}[width=\linewidth, title={}]
\textbf{Summary for RQ1:} Experimental results demonstrate that {\tool} achieves highly efficient model compression, reducing the model size to 0.63\% of the original, with a storage space of only 3 MB. Compared to the baseline model, the compression rate improves by 60\%.
\end{tcolorbox}

\subsection{RQ2: Can the {\tool} approach minimize the loss in accuracy for vulnerability assessment while compressing the model size?}
\label{sec:resultRQ2}

\textbf{Approach:} 
To assess whether the {\tool} approach can minimize accuracy loss in vulnerability assessment while compressing model size, we compare it with the original teacher model and the $\text{BiLSTM}_{soft}$ baseline. Following the evaluation metrics outlined in Section 4.4, we evaluate the accuracy metrics of the teacher model and both student models on the vulnerability assessment task to measure changes in model performance after knowledge distillation. The {\tool} approach uses the teacher model's output as soft labels during training, applying knowledge distillation to retain critical knowledge from the teacher model and aim for similar performance levels.

\textbf{Results:}
The performance comparison between {\tool} and $\text{BiLSTM}_{soft}$ is shown in Table~\ref{tab:rq1}. Experimental results demonstrate that {\tool} effectively maintains high accuracy while significantly reducing model size. Specifically, the teacher model achieves an accuracy of 60.93\% on the vulnerability assessment task. {\tool} reaches an accuracy of 54.39\%, which is 89.3\% of the teacher model's accuracy. In contrast, $\text{BiLSTM}_{soft}$ achieves an accuracy of 53.48\%, or 87.8\% of the teacher model’s performance, showing that {\tool} outperforms the $\text{BiLSTM}_{soft}$ approach. Compared to $\text{BiLSTM}_{soft}$, {\tool} reduces performance loss by 1.7\%, indicating that {\tool} minimizes accuracy loss while compressing the model.

\begin{tcolorbox}[width=1.0\linewidth, title={}]
\textbf{Summary for RQ2:} Experimental results show that {\tool} effectively preserves accuracy in the vulnerability assessment task while compressing model size. Specifically, the {\tool} approach retains 89.3\% of the teacher model's performance after compression, with an accuracy improvement of 1.7\% over the baseline model.
\end{tcolorbox}

\subsection{RQ3: To what extent does model size compression affect accuracy?}

\textbf{Approach:}
To examine the impact of model size compression on vulnerability assessment accuracy, we conduct comparative experiments using {\tool} models of different sizes. Specifically, we use student models compressed by {\tool} to sizes of 3 MB, 25 MB, and 50 MB, and compare them to the uncompressed teacher model CodeBERT (476 MB). We evaluate each model’s accuracy on the vulnerability assessment task and calculate the accuracy loss for each compressed model relative to the teacher model. This evaluation aims to quantify the effect of different compression levels on model performance, allowing us to analyze the relationship between compression efficiency and model accuracy.

\textbf{Results:} 
Table~\ref{tab:rq3} shows the accuracy of the three compressed models of different sizes on the vulnerability assessment task, as well as their performance loss relative to the teacher model. Results indicate a moderate decrease in accuracy as model size decreases. Specifically, the 3 MB {\tool} model achieves an accuracy of 54.39\%, reflecting a performance loss of 10.73\% compared to the teacher model. The 25 MB model achieves 54.55\% accuracy, with a performance loss of 10.47\%, while the 50 MB model reaches an accuracy of 56.21\%, reducing performance loss to 7.75\%. This suggests that larger compressed models (e.g., 50 MB) retain accuracy better, while smaller models (e.g., 3 MB) tend to experience higher accuracy loss.

\begin{table}[!]
    \centering
    \caption{Comparison of accuracy for compressed models of different sizes. The values in the ``Drop" row for the student models indicate their performance loss ratio relative to the teacher model.}
    \begin{tabular}{@{}ccccc@{}}
    \toprule
    \multirow{2}{*}{Approach} & \multicolumn{4}{c}{Vulnerability Assessment}   \\ 
    \cmidrule(l){2-5} 
     & Accuracy (\%) & & Drop (\%) & \\ \midrule
    CodeBERT (476 MB) & 60.93 &  & - & \\ \hdashline
    {\tool} (3 MB) & 54.39 &  & 10.73 & \\
    {\tool} (25 MB) & 54.55 &  & 10.47 & \\ 
    \bf{{\tool}} (50 MB) & \bf{56.21} &  & \bf{7.75} &  \\ \bottomrule
    \end{tabular}
    \label{tab:rq3}
    \end{table}

\begin{tcolorbox}[width=1.0\linewidth, title={}]
\textbf{Summary for RQ3:} Experimental results show that model size compression does indeed affect accuracy. Specifically, the 50 MB compressed model demonstrates only a 7.75\% performance decrease compared to the teacher model, while the 3 MB model exhibits a 10.73\% accuracy loss. However, given the substantial reduction in model size with a relatively small performance loss, the 3 MB model’s accuracy remains acceptable.
\end{tcolorbox}

\subsection{RQ4: What is the execution efficiency of the PSO algorithm?}

\textbf{Approach:}
To assess the execution efficiency of the PSO algorithm, we compare it with the Genetic Algorithm (GA), focusing on the average time cost during the search for the optimal model architecture. In the experiment, both PSO and GA methods are applied to search for the student model architecture. To ensure the reliability of the results, we independently run both the PSO and GA algorithms five times and record the time spent in each experiment. By calculating the average time cost from these five runs, we obtain the reduction ratio in time cost of the PSO method relative to the GA method.

\textbf{Results:} 
As shown in Table~\ref{tab:rq4}, during the search for the optimal student model architecture, the average time cost of GA was 0.43 seconds, while the average time cost of PSO was 0.28 seconds. Compared to GA, PSO reduced the time cost by 34.88\%. This result indicates that PSO demonstrates significantly higher execution efficiency and reduces the time overhead in the task of searching for the student model architecture. Upon further analysis of the algorithm execution principles, we find that PSO adjusts in each iteration by utilizing both global and local information of the particle swarm, while GA relies on the processes of reproduction and mutation, with a greater tendency for random mutation to explore the solution space. This may lead to longer convergence times, resulting in slightly lower efficiency in complex environments.

\begin{table}[!]
    \centering
    \caption{Comparison of the time cost between the PSO and GA approaches in searching for the optimal model architecture. The values in the "Drop" column represent the reduction ratio in the time cost of the PSO approach relative to the GA approach.}
    \begin{tabular}{@{}ccccc@{}}
    \toprule
    \multirow{2}{*}{Approach} & \multicolumn{4}{c}{Optimal Architecture Search}   \\ 
    \cmidrule(l){2-5} 
     & Average time cost (sec) & & Drop (\%) & \\ \midrule
    GA & 0.43 &  & - & \\
    \bf{PSO} & \bf{0.28} &  & \bf{34.88} & \\ 
    \bottomrule
    \end{tabular}
    \label{tab:rq4}
    \end{table}

\begin{tcolorbox}[width=1.0\linewidth, title={}]
\textbf{Summary for RQ4:} 
The execution efficiency of the PSO algorithm in model architecture search is clearly superior to that of the GA method. Compared to GA, PSO reduced the time cost by 34.88\%. The results suggest that PSO performs better in terms of execution efficiency.

\end{tcolorbox}

\section{Discussion}
\label{sec:Discussion}

\subsection{Time Cost of {\tool}}

The time consumption for model training has a significant impact on environmental sustainability. Therefore, we analyzed the differences in training time costs across compressed models of varying sizes, as shown in Table~\ref{tab:discuss1}. The original CodeBERT model requires 68 minutes of training time for the vulnerability assessment task, with a high time cost due to the large number of parameters and high computational complexity, resulting in slower inference.

However, our {\tool} approach uses particle swarm optimization to identify the optimal model architecture and applies knowledge distillation for model compression, effectively reducing training time costs. Specifically, models compressed to 3 MB, 25 MB, and 50 MB have training times of 19 minutes, 21 minutes, and 22 minutes, respectively—significantly lower than the 68 minutes required by CodeBERT. This result indicates that training time decreases substantially as model size is compressed. Notably, the 3 MB model reduces time cost by 72.1\%. As model size increases (from 3 MB to 50 MB), time costs rise slightly but remain within 22 minutes. This suggests that in selecting an appropriate model size, it is essential to balance accuracy with time costs to ensure high performance while maintaining low training time costs.

\begin{table}[!]
    \centering
    \caption{Comparison of time cost for compressed models of different sizes. The values in the ``Drop" row for the student models indicate the reduction in training time relative to the teacher model.}
    \begin{tabular}{@{}ccccc@{}}
    \toprule
    \multirow{2}{*}{Approach} & \multicolumn{4}{c}{Vulnerability Assessment}   \\ 
    \cmidrule(l){2-5} 
     & Time cost (min) & & Drop (\%) & \\ \midrule
    CodeBERT (476 MB) & 68 &  & - & \\ \hdashline
    \bf{{\tool} (3 MB)} & \bf{19} &  & \bf{72.1} & \\
    {\tool} (25 MB) & 21 &  & 69.1 & \\ 
    {\tool} (50 MB) & 22 &  & 67.6 &  \\ \bottomrule
    \end{tabular}
    \label{tab:discuss1}
    \end{table}

\subsection{Class Imbalance Problem}

In this section, we further analyze the impact of model compression on the MCC metric, as noted in Section~\ref{sec4.4}. The MCC metric assesses the model's ability to handle class imbalance issues. The results are shown in Table~\ref{tab:discuss2}. From the table, we observe that the original CodeBERT model achieves an MCC of 35.53\%, while our approach’s compressed student models—at 3 MB, 25 MB, and 50 MB—experience MCC reductions to 22.03\%, 23.72\%, and 26.74\%, corresponding to performance drops of 37.99\%, 33.24\%, and 24.74\%, respectively. Compared to the accuracy metric, MCC shows a greater degree of decline. For instance, in the 3 MB model, accuracy drops by 10.73\%, while MCC decreases by 37.99\%. This indicates that compressed models face considerable challenges in handling class imbalance, as compression reduces time and storage costs but compromises the model’s ability to address class imbalance issues effectively.

\begin{table}[!h]
    \centering
    \caption{Comparison of MCC metrics for compressed models of different sizes. The values in the ``Drop" row for the student models indicate the reduction in MCC relative to the teacher model.}
    \begin{tabular}{@{}ccccc@{}}
    \toprule
    \multirow{2}{*}{Approach} & \multicolumn{4}{c}{Vulnerability Assessment}   \\ 
    \cmidrule(l){2-5} 
     & MCC (\%) & & Drop (\%) & \\ \midrule
    CodeBERT (476 MB) & 35.53 &  & - & \\ \hdashline
    {\tool} (3 MB) & 22.03 &  & 37.99 & \\
    {\tool} (25 MB) & 23.72 &  & 33.24 & \\ 
    \bf{{\tool} (50 MB)} & \bf{26.74} &  & \bf{24.74} &  \\ \bottomrule
    \end{tabular}
    \label{tab:discuss2}
    \end{table}

\gcy{To further explore the robustness of our proposed approach {\tool} in handling the class imbalance issue, we conduct a separate inference and analysis on the minority class, specifically the low-frequency classes. As shown in Table~\ref{tab:sec:6.2}, CodeBERT achieves an accuracy of 51.53\% in vulnerability assessment for this type of class. $\text{BiLSTM}_{soft}$ has an accuracy of 34.52\%, with a drop of 33.01\% compared to CodeBERT. Our {\tool} approach, with the model size of only 3 MB, obtains an accuracy of 38.28\%, showing a drop of 25.71\% relative to CodeBERT. While the accuracy drop for {\tool} on the minority class is notable, it is important to note that this is a common issue among model compression methods. When reducing the model size, there is an inevitable loss of representational capacity, which can impact performance on minority classes. However, compared to $\text{BiLSTM}_{soft}$, our proposed approach {\tool} still demonstrates better performance. This indicates that, despite the challenges posed by model compression, {\tool} is relatively more effective in maintaining a certain level of accuracy on minority classes.}

\begin{table}[!]
    \centering
    \caption{\gcy{Performance Comparison of CodeBERT, $\text{BiLSTM}_{soft}$, and {\tool} on the Low-frequency Class (Low).}}
    \begin{tabular}{@{}ccccc@{}}
    \toprule
    \multirow{2}{*}{Approach} & \multicolumn{4}{c}{Vulnerability Assessment}   \\ 
    \cmidrule(l){2-5} 
     & Accuracy (\%) & & Drop (\%) & \\ \midrule
    CodeBERT (476 MB) & 51.53 &  & - & \\ \hdashline
    $\text{BiLSTM}_{soft}$ (7.5 MB) & 34.52 & & 33.01 \\
    {\tool} (3 MB) & 38.28 &  & 25.71 & \\
      \bottomrule
    \end{tabular}
    \label{tab:sec:6.2}
    \end{table}

\gcy{The performance degradation on minority classes after model compression can be attributed to the following reasons. First, during the compression process, some of the features that are crucial for identifying minority class samples may be pruned or not adequately preserved. Since minority classes have fewer samples, the model may not have sufficient data to learn and retain these distinctive features during compression. Second, the reduced model size limits the complexity of the model, making it more difficult to capture the subtle patterns that are characteristic of minority classes. This is especially true in the context of vulnerability assessment, where minority classes may represent rare or less-studied types of vulnerabilities.}

\gcy{To mitigate the loss in robustness after compression and improve performance on minority classes, the following potential solutions can be considered. One approach is to integrate class-aware loss functions. For example, focal loss can be used to assign higher weights to minority class samples during the training process. This encourages the model to pay more attention to these samples and improves its ability to classify them correctly. Another potential solution is to use data augmentation techniques specifically for minority samples. By increasing the number of minority class samples, the model can learn more robust features. Additionally, incorporating cost-sensitive learning during the distillation process can be beneficial. This means adjusting the cost associated with misclassifying minority class samples, so that the model is more penalized for incorrect predictions on these samples. We will consider these potential solutions in our future work and reduce the performance loss of the compressed model on minority classes.}

\gcy{\subsection{Generalization on Foundation Models}}

\gcy{In this section, we explore the generalization of {\tool} on VulBERTa~\cite{hanif2022vulberta}, a Transformer-based model commonly employed in vulnerability detection. Table~\ref{tab:sec:6.3} presents the comparison of model size and performance following the application of {\tool} for compressing the VulBERTa model.}

\begin{table*}[!h]
    \centering
    \caption{\gcy{Comparison of Model Size and Performance after Applying {\tool} to Compress VulBERTa model}
    }
    \normalfont
    \begin{tabular}{{ccccccc}}
    \toprule
    \multirow{2}{*}{Approach} & 
    \multicolumn{2}{c}{Capacity} &
    \multicolumn{2}{c}{Vulnerability Assessment} \\ 
    \cmidrule(l){2-7} 
     & Model Size (MB) & Drop (\%) & Accuracy (\%) & Drop (\%) & \\ \midrule
    \gcy{VulBERTa} & 499 & - & 60.51 & - \\
    \hdashline     
    \textbf{{\tool}} & \textbf{3} & \textbf{99.4} & \bf{53.89} &  \bf{10.94}  \\
    
    \bottomrule
    \end{tabular}
    \label{tab:sec:6.3}
\end{table*}

\gcy{
The VulBERTa model initially has a model size of 499 MB and an accuracy of 60.51\% in vulnerability assessment. After applying {\tool}, the model size is significantly reduced to 3 MB, resulting in a compression rate of 99.4\%. Although there is a 10.94\% decrease in accuracy, leading to a value of 53.89\%, the substantial reduction in model size highlights the effectiveness of the {\tool} approach. This result not only validates the generalizability of {\tool} across different architectures, particularly those based on the Transformer, but also underscores its potential in resource-constrained scenarios where a minimized model size is crucial.}

\gcy{\subsection{Generalization across Programming Languages}}

\gcy{To further validate the generalizability of our proposed approach {\tool} across different programming languages, we conduct additional experiments on the MegaVulJava dataset, which contains Java vulnerabilities in MegaVul~\cite{ni2024megavul} by following the same experimental setup. Table~\ref{tab:sec:6.4} presents the performance comparison of CodeBERT, $\text{BiLSTM}_{soft}$, and {\tool} on this Java-based dataset.}

\gcy{CodeBERT, with a model size of 476 MB, achieves an accuracy of 62.45\% in vulnerability assessment. $\text{BiLSTM}_{soft}$, having a model size of 7.5 MB, shows an accuracy of 54.26\%, with a drop of 13.11\% compared to CodeBERT. Our proposed approach {\tool}, despite having a much smaller model size of only 3 MB, still managed to attain an accuracy of 57.69\%, with a relatively lower drop of 7.62\% compared to CodeBERT. These results indicate that {\tool} can maintain a reasonable level of performance even when applied to vulnerability assessment in the Java programming language, demonstrating its generalization across different programming languages.}

\begin{table}[!]
    \centering
    \caption{\gcy{Performance Comparison of CodeBERT, $\text{BiLSTM}_{soft}$, and {\tool} on the MegaVulJava dataset.}}
    \begin{tabular}{@{}ccccc@{}}
    \toprule
    \multirow{2}{*}{Approach} & \multicolumn{4}{c}{Vulnerability Assessment}   \\ 
    \cmidrule(l){2-5} 
     & Accuracy (\%) & & Drop (\%) & \\ \midrule
    CodeBERT (476 MB) & 62.45 &  & - & \\ \hdashline
    $\text{BiLSTM}_{soft}$ (7.5 MB) & 54.26 & & 13.11 \\
    {\tool} (3 MB) & 57.69 &  & 7.62 & \\
      \bottomrule
    \end{tabular}
    \label{tab:sec:6.4}
    \end{table}

\subsection{\gcy{Analysis of Hyperparameter Value Settings}}

\gcy{To ensure the effectiveness and efficiency of our proposed approach, we systematically analyze the impact of key hyperparameters on both the PSO optimization algorithm and the model training process. By independently varying each parameter while keeping others fixed, we can observe their individual effects on search performance, convergence behavior, and overall model accuracy. This analysis can not only help to identify optimal parameter settings for our specific task but also provide valuable insights and practical guidelines for parameter selection in similar applications. }

\subsubsection{\gcy{Analysis of PSO Parameters}}

\gcy{The performance of the PSO algorithm hinges on key parameters like population size, number of iterations, and inertia weight. Here, we meticulously analyze these parameters and their implications for the optimization process.}

\begin{figure*}[t]
    \centering
    \begin{subfigure}[b]{0.32\textwidth}
        \centering
        \includegraphics[width=\textwidth]{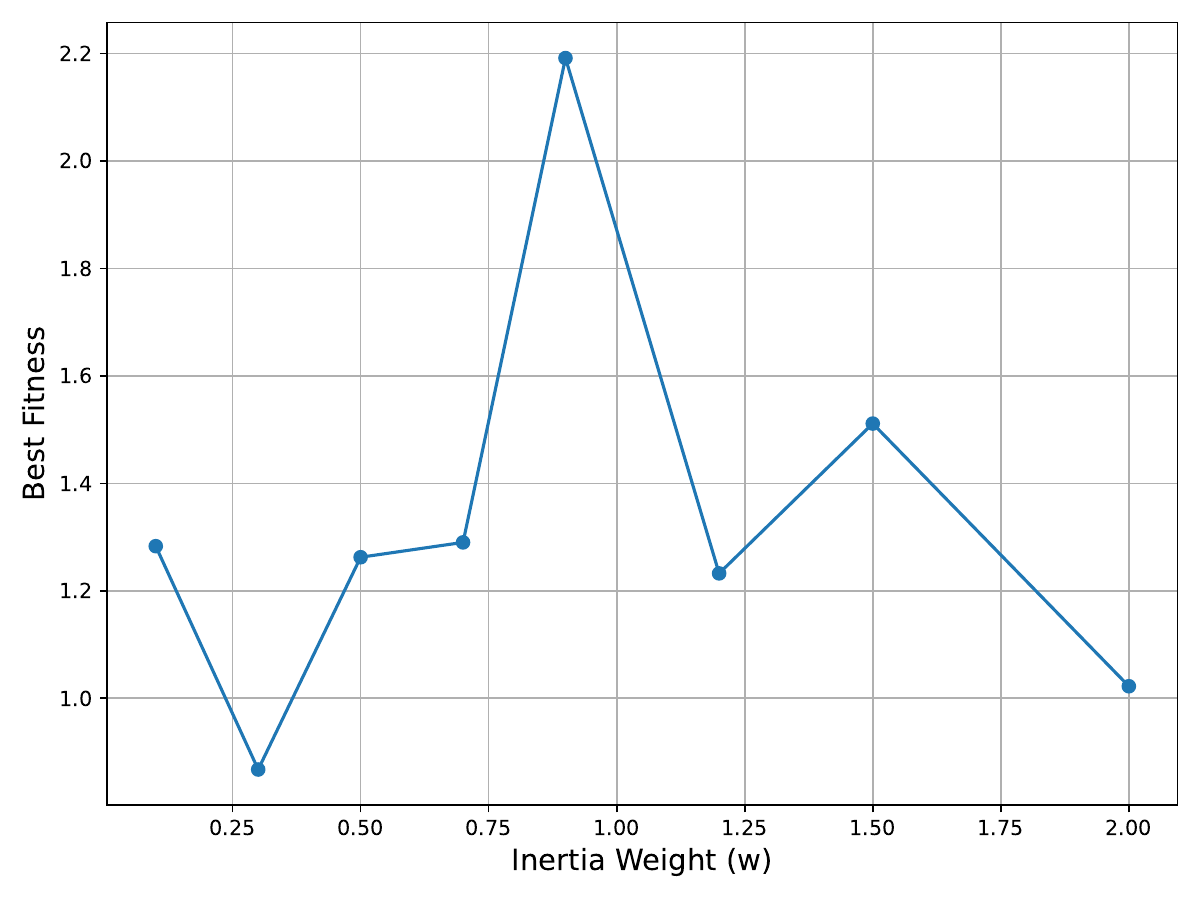}
        \caption{Inertia Weight (w)}
        \label{fig:Inertia Weight}
    \end{subfigure}
    \hfill
    \begin{subfigure}[b]{0.32\textwidth}
        \centering
        \includegraphics[width=\textwidth]{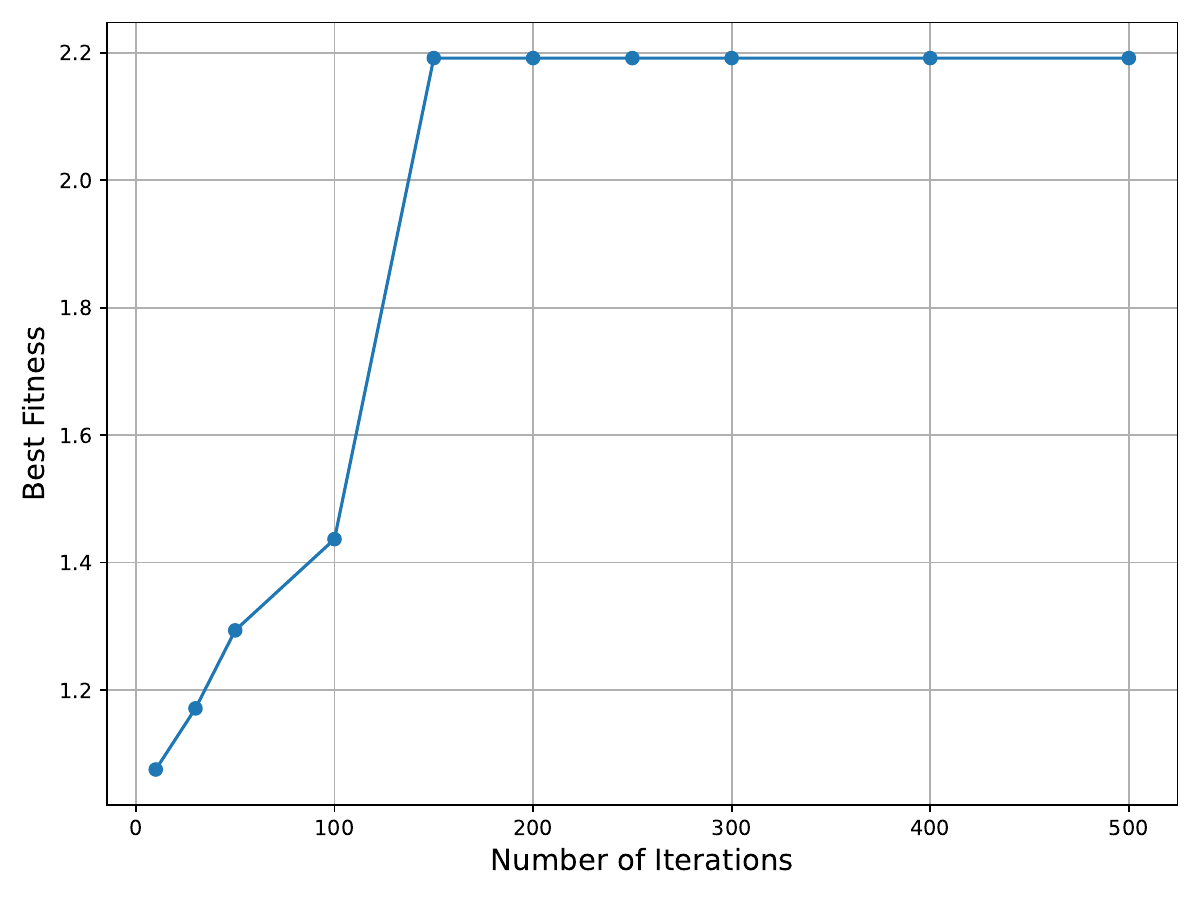}
        \caption{Number of Iterations}
        \label{fig:Number of Iterations}
    \end{subfigure}
    \hfill
    \begin{subfigure}[b]{0.32\textwidth}
        \centering
        \includegraphics[width=\textwidth]{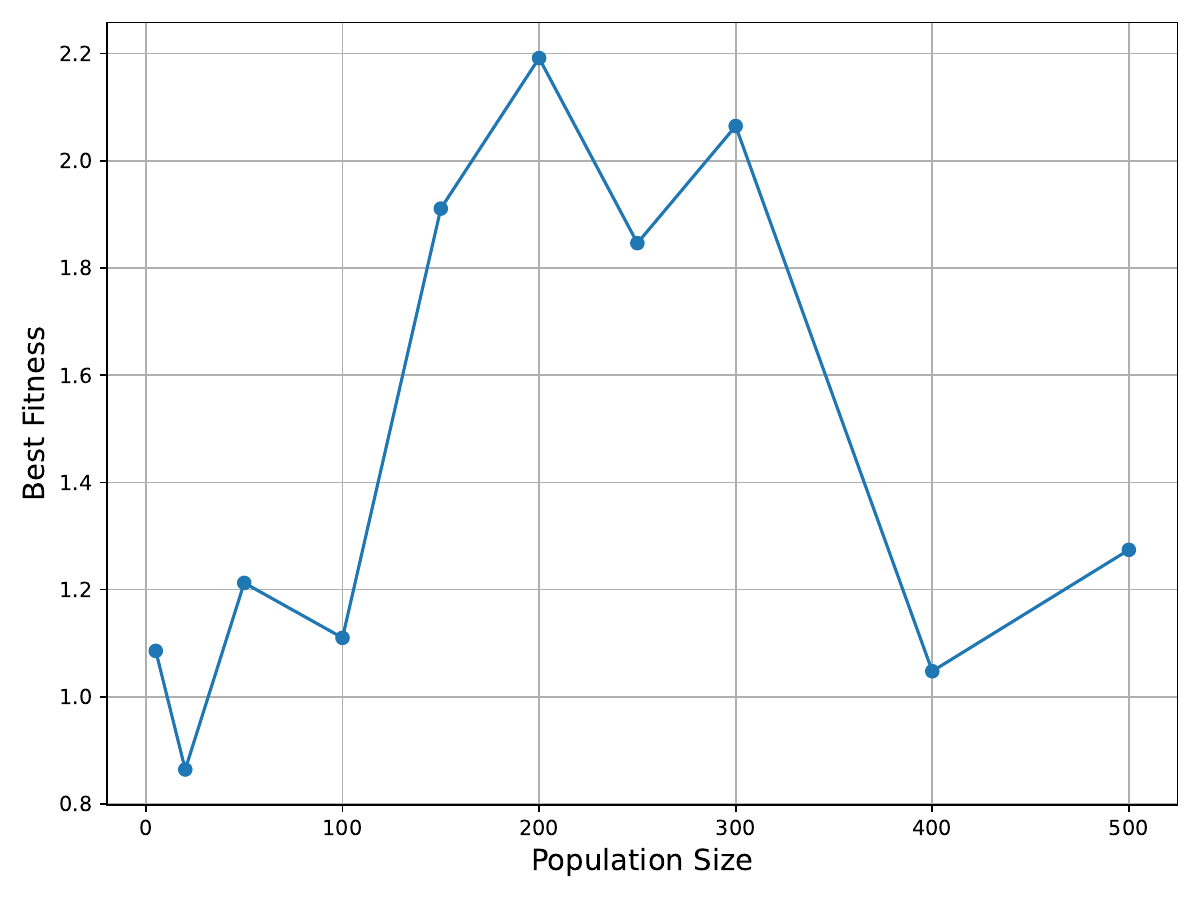}
        \caption{Population Size}
        \label{fig:Population Size}
    \end{subfigure}
    \caption{\gcy{Results of different parameter values of PSO  in {\tool}.}}
    \label{fig:pso_para}
\end{figure*}

\gcy{Regarding the inertia weight, as shown in Figure~\ref{fig:Inertia Weight}, we hold the population size and number of iterations at their optimal levels determined through prior experimentation. By incrementally adjusting the inertia weight, we observe a non-linear relationship with the best fitness value. As the inertia weight increases from a low value, the best fitness value decreases, indicating that the algorithm explores different regions of the search space. When the inertia weight reaches 0.9, the best fitness value reaches its minimum, demonstrating that the algorithm effectively exploits a promising area. However, as the inertia weight increases beyond 0.9, the best fitness value begins to rise again, revealing that the balance between exploration and exploitation is disrupted. This behavior shows that the algorithm achieves the optimal fitness value when the inertia weight is set to 0.9, representing the convergence point for balancing exploration and exploitation.}

\gcy{For the number of iterations, with the population size and inertia weight maintained at their respective optimal settings, we observe the evolution of the best fitness value. As shown in Figure~\ref{fig:Number of Iterations}, in the initial stages (0 to 150 iterations), the best fitness value improves rapidly, indicating that the algorithm makes significant progress in finding better solutions. After 150 iterations, the best fitness value stabilizes at 2.2, which is a clear sign of convergence. This result demonstrates that the algorithm reaches a point where additional iterations yield no further improvements. Therefore, the PSO algorithm converges effectively within 150 iterations for our vulnerability assessment model compression task.}

\gcy{Concerning the population size, we fix the inertia weight and number of iterations at their optimal values. As illustrated in Figure~\ref{fig:Population Size}, increasing the population size initially causes the best fitness value to fluctuate, then rise significantly at a population size of 100, and reach its maximum at a population size of 200. When the population size exceeds 200, the best fitness value begins to decline, indicating that although a larger population can help to enhance exploration, it also results in more search costs. This result demonstrates that the algorithm achieves optimal performance with a population size of 200, effectively balancing search space exploration and computational efficiency.}

\gcy{Overall, this in-depth convergence analysis validates the reliability of our PSO-based optimization process and provides guidance for parameter selection in future applications. By understanding how these parameters influence convergence, we can fine-tune the PSO algorithm to achieve more efficient and accurate optimization results in vulnerability assessment model compression.}

\subsubsection{\gcy{Analysis of Model Training Parameters}}

\begin{figure*}[t]
    \centering
    \begin{subfigure}[b]{0.49\textwidth}
        \centering
        \includegraphics[width=\textwidth]{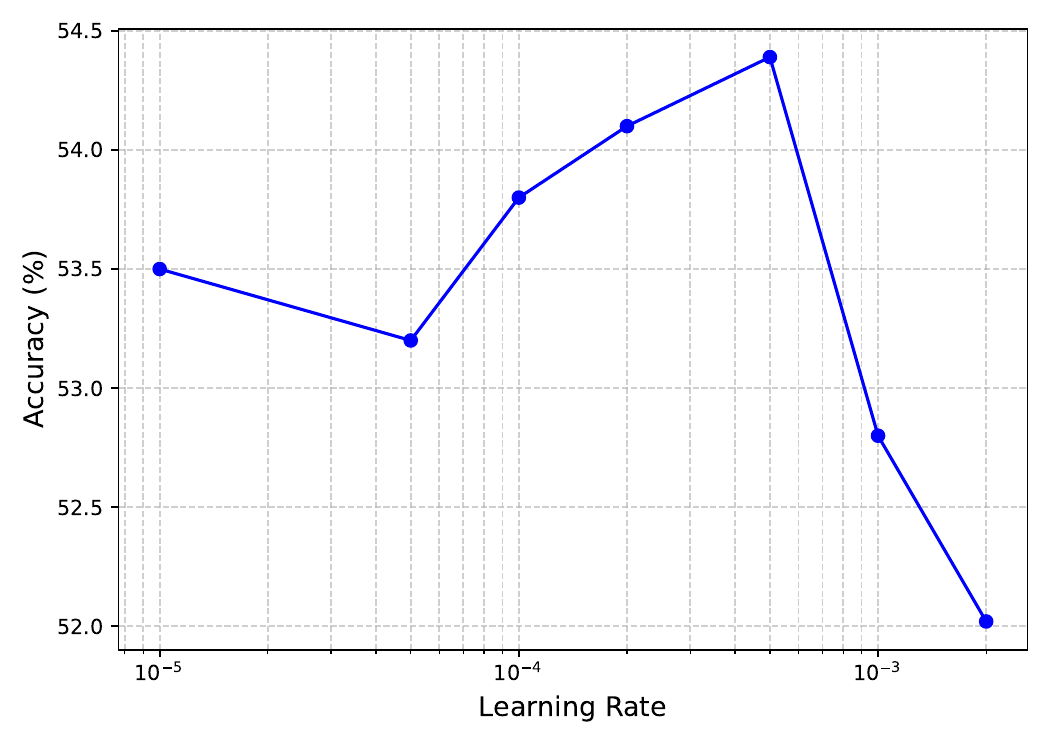}
        \caption{Learning Rate}
        \label{fig:learning_rate}
    \end{subfigure}
    \hfill
    \begin{subfigure}[b]{0.49\textwidth}
        \centering
        \includegraphics[width=\textwidth]{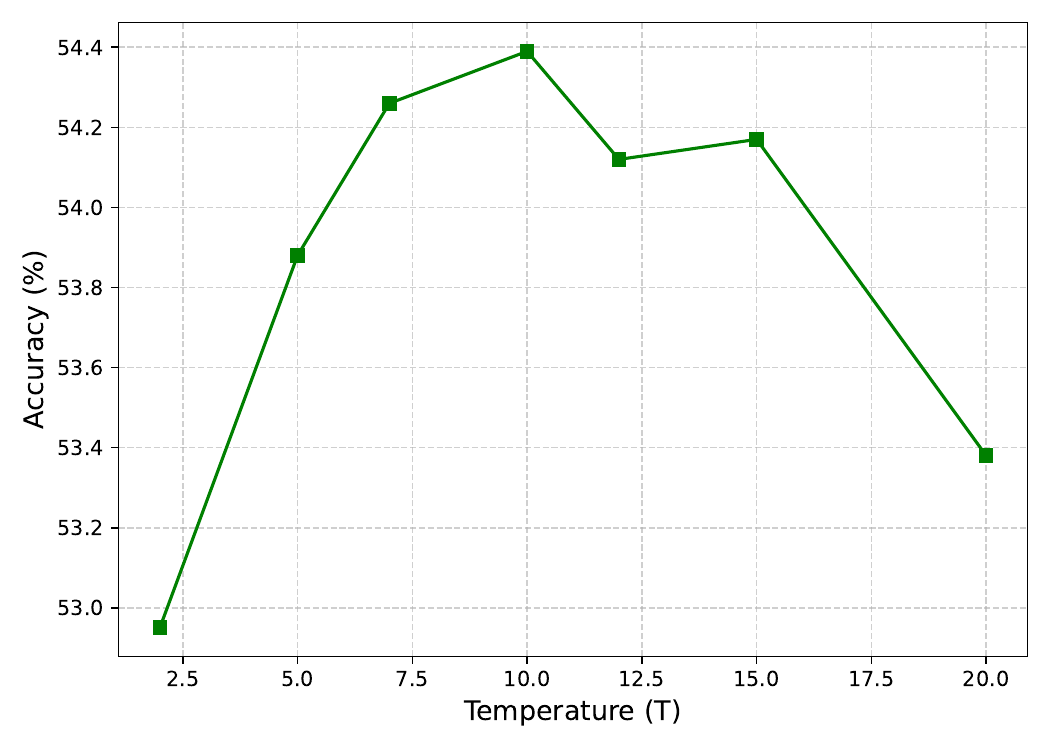}
        \caption{Temperature (T)}
        \label{fig:temperature}
    \end{subfigure}
    \caption{\gcy{Results of model training parameters.}}
    \label{fig:train_para}
\end{figure*}

\gcy{The learning rate is a crucial hyperparameter that dictates the step size at which the model updates its weights during training. In our study, we recognize the importance of determining an appropriate learning rate for the student model, considering its architecture compared to the teacher model. To this end, we conduct a sensitivity analysis, varying the learning rate across a range of values: 1e-5, 5e-5, 1e-4, 2e-4, 5e-4, 1e-3, 2e-3.}

\gcy{As illustrated in Figure~\ref{fig:learning_rate}, the accuracy of the student model demonstrates a non-linear relationship with the learning rate. Starting from a learning rate of 1e-5, the accuracy is 53.5\%. As the learning rate decreases to 5e-5, the accuracy dips slightly to approximately 53.2\%. When the learning rate increases to 1e-4, the accuracy begins to rise, reaching around 53.8\%. Further increasing the learning rate to 5e-4 leads to a peak accuracy of about 54.39\%. However, as the learning rate continues to increase to 1e-3, the accuracy drops significantly to around 52.8\%. This analysis reveals that a very low learning rate may cause the model to converge slowly, potentially getting trapped in sub-optimal solutions. On the other hand, a learning rate that is too high can lead to overshooting the optimal solution during weight updates, resulting in unstable training and a decrease in performance. Based on these results, we can find that an intermediate learning rate value, such as 5e-4, yields relatively better performance for the student model. This provides a rational basis for our choice of learning rate.}

\gcy{The temperature parameter ($T$) plays a critical role in the knowledge distillation process by controlling the smoothness of the teacher model's output distribution. To explore its impact on the student model's performance, we perform a sensitivity analysis by evaluating the student model across different temperature values: 2, 5, 7, 10, 12, 15, and 20.}

\gcy{As shown in Figure~\ref{fig:temperature}, when the temperature is set at 2, the accuracy of the student model is relatively low. As the temperature increases to 5, the accuracy rises to 53.88\%. The accuracy continues to increase, peaking at 54.39\% when the temperature reaches 10. However, as the temperature further increases to 12, the accuracy slightly decreases to 54.12\%. As the temperature increases beyond 15, the accuracy starts to decline more notably. Low values of T result in a more peaked soft label distribution. This may underutilize the relational information between classes, as the probabilities of non-dominant classes are suppressed too much, preventing the student model from learning subtle relationships. Conversely, higher values of $T$ overly smooth the distribution. In this case, the differences between classes become less distinct, causing the student model to focus less on the correct class. An intermediate value, such as $T$ = 10, achieves the best balance. It allows the student model to benefit most from the softened logits while still maintaining class distinction, thus optimizing the knowledge transfer process and enhancing the overall performance of the student model.}

\subsection{\gcy{Efficiency Analysis of {\tool}}}

\gcy{In practical deployment scenarios, the efficiency of a vulnerability assessment model is of great significance. In this subsection, we conduct a comprehensive efficiency analysis of {\tool}, focusing on inference latency and computational cost metrics, such as inference latency and GFLOPs.}

\begin{table}[!]
    \centering
    \caption{\gcy{Comprehensive efficiency analysis of {\tool}, here ``Latency" denotes the inference latency.}}
    \begin{tabular}{@{}ccccc@{}}
    \toprule
    \multirow{2}{*}{Approach} & \multicolumn{4}{c}{Vulnerability Assessment}   \\ 
    \cmidrule(l){2-5} 
     & Latency (ms) & & GFLOPs & \\ \midrule
    CodeBERT (476 MB) & 1635 &  & 86.54 & \\ \hdashline
    $\text{BiLSTM}_{soft}$ (7.5 MB) & 189 & & 13.39 \\
    {\tool} (3 MB) & \textbf{58} &  & \textbf{3.82} & \\
      \bottomrule
    \end{tabular}
    \label{tab:sec:6.6}
    \end{table}

\gcy{Inference latency is a crucial metric for evaluating the real-time performance of a model. As presented in Table~\ref{tab:sec:6.6}, CodeBERT, with a model size of 476 MB, has an inference latency of 1635 ms. $\text{BiLSTM}_{soft}$, with a model size of 7.5 MB. Our proposed approach {\tool}, with a significantly smaller model size of only 3 MB, further reduces the inference latency to 58 ms, representing a remarkable reduction of about 96.4\% compared to CodeBERT. This substantial reduction in latency indicates that {\tool} can enable faster vulnerability assessments, which is highly desirable in practical applications where quick responses are required. For instance, in real-time security monitoring systems, a lower latency allows for more timely detection and response to potential vulnerabilities.}

\gcy{GFLOPs is used to measure the computational cost of a model. CodeBERT has a relatively high GFLOPs value of 86.54, reflecting its high computational complexity. $\text{BiLSTM}_{soft}$ reduces this to 13.39 GFLOPs. {\tool} achieves an even lower GFLOPs value of 3.82, representing a significant reduction of about 95.6\% compared to CodeBERT. A lower GFLOPs value implies that {\tool} requires fewer computational resources during inference, which not only reduces the processing time but also potentially lowers energy consumption. This makes {\tool} more suitable for deployment in resource-constrained environments.
In summary, the efficiency analysis of {\tool} demonstrates its superiority in terms of both inference speed and computational resource utilization. These results highlight the practical viability of {\tool} for real-world vulnerability assessment tasks, where efficiency is often a critical factor.}

\subsection{Threats to Validity}
In this subsection, we discuss the potential threats to the validity of our study.

\textbf{Internal Threats}. 
Our experimental results may be influenced by the experimental setup and model selection. In implementing the {\tool} approach, we used consistent hyperparameter settings to minimize the impact of different configurations. However, certain hyperparameter choices in the knowledge distillation process (such as temperature parameters and loss function settings) could affect the student model’s performance. To ensure consistency in results, please use the parameters provided for model training. Additionally, using unlabeled data for distillation may impact model performance; future studies could explore using different datasets for further validation.

\textbf{External Threats}. 
The external validity of this study is primarily limited by the representativeness of the dataset and task chosen. We used an enhanced version of the MegaVul dataset for vulnerability assessment, which includes only C++ code. This may limit the generalizability of the {\tool} approach to other programming languages or tasks in different domains. To improve applicability, we plan to test additional datasets and tasks in future research to verify the model’s performance across a broader range of scenarios.

\textbf{Construct Threats:} 
Our study uses model size, time cost, accuracy, and MCC as evaluation metrics, with MCC reflecting class imbalance issues. However, the interpretation of these metrics may be affected by model compression. For example, the significant decrease in MCC after compression indicates that the student model may struggle with class imbalance. Future research could explore additional evaluation metrics to more comprehensively assess the impact of model compression on performance.

\textbf{Conclusion Threats:} 
Our experimental conclusions are based on a specific model and dataset and may not directly generalize to other models or application scenarios. Although we enhance the robustness of our conclusions through multiple experiments and comparison with baseline approaches, the performance of the {\tool} approach may vary in different hardware environments and dataset sizes. Future research should validate these findings across diverse experimental settings and platforms to ensure the generality and reliability of the conclusions.

\section{Related Work}
\label{sec:Related Work}

\textbf{Software vulnerability assessment (SVA).} Existing vulnerability assessment methods, which primarily rely on textual descriptions of vulnerabilities or source code analysis, have notable limitations. Han et al.~\cite{han2017learning} proposed using word embeddings and shallow convolutional neural networks to extract discriminative features from vulnerability descriptions for severity prediction; however, their model's expressive capability is limited. Spanos et al.~\cite{spanos2018multi} developed a model that combines text analysis with multi-objective classification to estimate vulnerability characteristics and calculate severity scores, but the computational cost is high. Liu et al.~\cite{liu2019vulnerability} introduced deep learning techniques for vulnerability text classification, achieving performance gains at the expense of significant training time, making it less efficient. Le et al.~\cite{le2019automated} systematically combined character and word features, yet their approach requires extensive hyperparameter tuning, adding complexity to model deployment. Babalau et al.~\cite{babalau2021severity} applied a pre-trained BERT model within a multi-task learning framework for vectorizing vulnerability descriptions but still required additional experiments to optimize hyperparameters and loss weights.

Source code-based analysis methods also have their shortcomings. Ganesh et al.~\cite{ganesh2021predicting} evaluated the effectiveness of machine learning models in predicting vulnerabilities from source code but found that these models performed poorly in making accurate predictions. Le et al.~\cite{le2022use} proposed a function-level vulnerability assessment approach, using vulnerable code statements as input for model development; however, the model’s generalization capabilities were limited. Hao et al.~\cite{hao2023novel} introduced a rapid assessment method using function call graphs and vulnerability attribute graphs that bypasses the need for vulnerability reports or manual analysis, but the model's complexity may hinder practical applications. Compared to our vulnerability assessment approach based on model distillation, these methods suffer from high computational costs, complex tuning requirements, or lower efficiency.

\textbf{Model compression and knowledge distillation.} Mukherjee et al.~\cite{mukherjee2019cogni}proposed leveraging knowledge distillation to transfer visual discriminative knowledge from a pre-trained image classification model to a deep bidirectional long short-term memory network operating on EEG signals, achieving state-of-the-art performance in brain signal classification.
Peng et al.~\cite{peng2019few} introduced a Knowledge Transfer Network (KTN) that utilizes prior knowledge from base classes to assist in recognizing novel classes with limited examples, effectively improving recognition accuracy in few-shot scenarios. Pan et al.~\cite{pan2019novel} presented an enhanced collaborative autoencoder model incorporating knowledge distillation to improve recommendation performance, demonstrating superior quality in top-\emph{n} recommendations compared to traditional collaborative filtering methods. 
Chen et al.~\cite{chen2018adversarial} proposed an adversarial distillation framework that integrates external knowledge into recommendation systems, leading to more efficient and accurate recommendations through adversarial training between teacher and student models.
Bai et al.~\cite{bai2019learn} explored transferring knowledge from pre-trained language models to sequence-to-sequence speech recognition systems, enhancing spelling accuracy by aligning the output distributions of the teacher language model and the student speech recognition model.
Ng et al.~\cite{ng2018teacher} introduced a teacher-student training framework for text-independent speaker recognition, where the teacher model guides the student model to learn speaker-discriminative features, resulting in improved performance in speaker recognition tasks.
Wang et al.~\cite{wang2020minilm} presented MiniLM, a model that compresses large pre-trained transformers by distilling deep self-attention knowledge, retaining performance while significantly reducing model size and computational requirements.
Liang et al.~\cite{liang2023homodistil} introduced HomoDistil, a distillation method that employs homotopic transformations to distill knowledge from large pre-trained transformers into smaller models, maintaining task-agnostic properties to ensure applicability across diverse tasks.
Li et al.~\cite{li2020knowledge} proposed a knowledge distillation framework tailored for multi-task learning scenarios, enabling a single student model to perform multiple tasks effectively by distilling knowledge from multiple task-specific teacher models.
\gcy{Zhou et al.~\cite{zhou2023wavenet} proposed WaveNet, a wavelet-based MLP architecture for salient object detection in RGB-thermal infrared images, which employs knowledge distillation from a transformer teacher and utilizes discrete wavelet transforms for cross-modality feature fusion, achieving impressive results on benchmark datasets.} \gcy{Building upon this research, Zhou et al~\cite{zhou2024hybrid} proposed HKDNet, a lightweight hybrid knowledge distillation network for RGB-thermal crowd density estimation, which efficiently combines convolution and self-attention while significantly reducing computational cost and model size compared to a heavyweight teacher network.} \gcy{Continuing the exploration of multimodal learning, Zhou et al.~\cite{zhou2025knowledge} proposed SSRNet-S*, a multimodal transmission-line detection network for RGB-thermal images that integrates knowledge distillation and contrastive learning to enhance feature representation, reduce model size, and improve robustness under challenging conditions, achieving superior performance with significantly fewer parameters.} \gcy{In addition to these efforts, Zhou et al.~\cite{zhou2025feature} proposed FCDENet for RGB-D indoor scene classification, introducing feature contrast difference and information clustering modules, as well as wavelet-based cross-layer decoding, to effectively enhance feature representation and achieve superior accuracy on benchmark indoor datasets.}
These studies collectively demonstrate the versatility and effectiveness of knowledge distillation across various domains, including cognitive feature learning, image recognition, recommendation systems, speech recognition, speaker recognition, model compression, and multi-task learning.

\section{Conclusion and Future Work}
\label{sec:Conclusion}

In this study, we propose the {\tool} method, which leverages a combination of particle swarm optimization and knowledge distillation techniques to construct a lightweight software vulnerability assessment model. The {\tool} method effectively compresses the model size while retaining much of the original model’s performance. Experimental results demonstrate that compared to the original CodeBERT model, {\tool} significantly reduces storage requirements and inference time while achieving competitive performance on accuracy and other metrics.

Despite the promising performance of {\tool}, there are several areas for further research. First, future work could explore more advanced distillation strategies to enhance model performance on imbalanced datasets. Second, it would be valuable to validate the applicability of {\tool} across different programming languages and task scenarios to improve its generalizability. Finally, combining other model compression techniques, such as quantization and pruning, could further optimize the model’s inference speed and performance.

\section*{CRediT authorship contribution statement}

\textbf{Chaoyang Gao:} Data curation, Software, Validation, Conceptualization, Methodology, Writing -review \& editing.
\textbf{Xiang Chen:} Conceptualization, Methodology, Writing -review \& editing, Supervision.
\textbf{Jiyu Wang}: Conceptualization, Data curation, Software.
\textbf{Jibin Wang}: Conceptualization, Data curation, Software.
\textbf{Guang Yang}: Conceptualization, Data curation, Software.

\section*{Declaration of competing interest}
The authors declare that they have no known competing financial interests or personal relationships that could have appeared to
influence the work reported in this paper.

\section*{Data availability}
Data will be made available on request.

\section*{Acknowledgments}
\gcy{
The authors would like to thank the editors and the anonymous reviewers for their insightful comments and suggestions, which can substantially improve the quality of this work. 
Chaoyang Gao and Xiang Chen have contributed equally to this work and are co-first authors. Xiang Chen is the corresponding author. 
This research was partially supported by the National Natural Science Foundation of China (Grant no. 61202006), the Open Project of State Key Laboratory for Novel Software Technology at Nanjing University under (Grant No. KFKT2024B21) and the Postgraduate Research \& Practice Innovation Program of Jiangsu Province (Grant nos. SJCX24\_2022).
}

\bibliography{mylib}
\bibliographystyle{elsarticle}

\vspace{1cm}

\noindent\textbf{Chaoyang Gao} 
is currently pursuing the Master degree at the School of Artificial Intelligence and Computer Science, Nantong University. His research interests include software
repository mining.
\par
\vspace{1cm}

\noindent\textbf{Xiang Chen} 
 received the B.Sc. degree in the school of management from Xi'an Jiaotong University, China in 2002. Then he received his M.Sc., and Ph.D. degrees in computer software and theory from Nanjing University, China in 2008 and 2011 respectively. 
He is currently an Associate Professor at the School of Artificial Intelligence and Computer Science, Nantong University. He has authored or co-authored more than 120 papers in refereed journals or conferences, such as IEEE Transactions on Software Engineering, ACM Transactions on Software Engineering and Methodology, Empirical Software Engineering, Computer \& Security, Software Testing, Verification and Reliability, Information and Software Technology, Journal of Systems and Software, IEEE Transactions on Reliability, Journal of Software: Evolution and Process, Software - Practice and Experience, Automated Software Engineering, International Conference on Software Engineering (ICSE), International Conference on the Foundations of Software Engineering (FSE), International Conference Automated Software Engineering (ASE), International Conference on Software Maintenance and Evolution (ICSME), International Conference on Program Comprehension (ICPC), and International Conference on Software Analysis, Evolution and Reengineering (SANER). His research interests include software engineering, in particular software testing and maintenance, software repository mining, and empirical software engineering. He received two ACM SIGSOFT distinguished paper awards in ICSE 2021 and ICPC 2023. He is the editorial board member of Information and Software Technology. More information about him can be found at: 
\url{https://xchencs.github.io/index.html}.

\par
\vspace{1cm}

\noindent\textbf{Jiyu Wang} 
is currently pursuing the Master degree at the School of Artificial Intelligence and Computer Science, Nantong University. His research interests include vulnerability detection and repair.

\par
\vspace{1cm}

\noindent\textbf{Jibin Wang} 
is currently pursuing the Bachelor degree at the School of Artificial Intelligence and Computer Science, Nantong University. Her research interests include automatic vulnerability assessment.

\par
\vspace{1cm}

\noindent\textbf{Guang Yang} 
received the M.D. degree in computer technology from Nantong University, Nantong, in 2022. Then he is currently pursuing the Ph.D degree at Nanjing University of Aeronautics and Astronautics, Nanjing.
His research interest is AI4SE and he has authored or co-authored more than 30 papers in refereed journals or conferences, such as IEEE Transactions on Software Engineering(TSE), ACM Transactions on Software Engineering and Methodology (TOSEM), Empirical Software Engineering (EMSE), Journal of Systems and Software (JSS), International Conference on Software Maintenance and Evolution (ICSME), and International Conference on Software Analysis, Evolution and Reengineering (SANER).
More information about him can be found at: \url{https://ntdxyg.github.io/}.

\end{document}